\documentclass[10pt,twocolumn,twoside]{IEEEtran}

\usepackage{times}
\usepackage{url}
\usepackage[pdftex]{graphics,graphicx}
\usepackage[caption=false,font=footnotesize]{subfig}
\usepackage{placeins}
\usepackage{fixltx2e}
\usepackage{cite}
\usepackage{algorithm}
\usepackage{algorithmic}
\usepackage[cmex10]{amsmath}
\usepackage{amssymb}
\usepackage{amsthm}
\usepackage{todonotes}
\usepackage{booktabs}
\usepackage{array}
\usepackage{color}

\newcommand{\nystrom}{Nystr\"{o}m}
\newcommand{\wtg}{\widetilde G}
\newcommand{\kp}{_{k+1}}
\newcommand{\inv}{^{-1}}
\newcommand{\dinv}{^{\dagger}}
\newcommand{\eqb}[1]{\begin{equation}\label{#1}}
\newcommand{\eqe}{\end{equation}}
\newcommand{\Lam}{\Lambda}
\newcommand{\nLam}{{\Lambda^{c}}}

\DeclareMathOperator*{\argmax}{arg\,max}
\DeclareMathOperator*{\colsum}{colsum}
\DeclareMathOperator*{\diag}{diag}

\newtheorem{THEO}{Theorem}
\newtheorem{LEMM}{Lemma}

\begin{document}
\bstctlcite{IEEE:BSTcontrol}

\title{oASIS:  Adaptive Column Sampling \\ for Kernel Matrix Approximation}

\author{Raajen~Patel*,~\IEEEmembership{Student Member,~IEEE,}
        Thomas~A.~Goldstein,~\IEEEmembership{Member,~IEEE,}
                Eva~L.~Dyer,~\IEEEmembership{Member,~IEEE,}
                Azalia~Mirhoseini,~\IEEEmembership{Student Member,~IEEE,}
                and~Richard~G.~Baraniuk,~\IEEEmembership{Fellow,~IEEE}
\thanks{R. Patel, A. Mirhoseini, and R. Baraniuk are with the Department of Electrical and Computer Engineering, Rice University, Houston, Texas USA. e-mail: rjp2@rice.edu.}%
\thanks{T. A. Goldstein is with the Department of Computer Science, University of Maryland, College Park, Maryland USA.}%
\thanks{E. L. Dyer is with the Department of Physical Medicine and Rehabilitation, Rehabilitation Institute of Chicago, Northwestern University, Chicago, Illinois USA.}%
\thanks{This work was supported in part by grants NSF CCF-0926127, NSF CCF-1117939, ONR N00014-12-1-0579, ONR N00014-11-1-0714, and
ARO MURI W911NF-09-1-0383, and by the Data Analysis and Visualization Cyberinfrastructure funded by NSF under grant OCI-0959097.}}

\markboth{Patel \MakeLowercase{\textit{et al.}}: oASIS:  Adaptive Column Sampling for Kernel Matrix Approximation}%
{Patel \MakeLowercase{\textit{et al.}}: oASIS:  Adaptive Column Sampling for Kernel Matrix Approximation}

\maketitle

\begin{abstract}
Kernel matrices (e.g. Gram or similarity matrices) are essential for many state-of-the-art approaches to classification, clustering, and dimensionality reduction. For large datasets, the cost of forming and factoring such kernel matrices becomes intractable. To address this challenge, we introduce a new adaptive sampling algorithm called Accelerated Sequential Incoherence Selection (oASIS) that samples columns without explicitly computing the entire kernel matrix. We provide conditions under which oASIS is guaranteed to exactly recover the kernel matrix with an optimal number of columns selected. Numerical experiments on both synthetic and real-world datasets demonstrate that oASIS achieves performance comparable to state-of-the-art adaptive sampling methods at a fraction of the computational cost. The low runtime complexity of oASIS and its low memory footprint enable the solution of large problems that are simply intractable using other adaptive methods.
\end{abstract}


\section{Introduction}

\subsection{Kernel Matrix Approximation}
\label{ss:kma}

\IEEEPARstart{M}{any} machine learning and data analysis frameworks for classification, clustering, and dimensionality reduction require the formation of kernel matrices that contain the pairwise ``similarities'' or distances between signals in a collection of data. For instance, kernel methods for classification \cite{Williams01usingthe}, nonlinear dimensionality reduction \cite{Coifman20065, tenenbaum_global_2000}, and spectral clustering \cite{Shi:2000:NCI:351581.351611} all require computing and storing an $n \times n$ kernel matrix, where $n$ is the number of examples (points) in the dataset.

As the size $n$ of the dataset grows, the computation and storage of kernel matrices becomes increasingly difficult. For instance, explicitly storing a kernel matrix with dimension $n=10^5$ using IEEE standard binary64 requires $80$ gigabytes of memory \cite{4610935}. To extend kernel-based methods to extremely large $n$, a host of methods have focused on sampling a small subset of columns from the kernel matrix and then computing a low-rank approximation of the matrix using some flavor of the {\em \nystrom{} method} \cite{journals/jmlr/DrineasM05}. 

An accurate low-rank approximation determines and keeps only the most important dimensions in the column space of the matrix. In doing so, it captures the majority of the information in a matrix with a smaller ambient dimensional space. In the kernel matrix setting, the size of the dataset is larger than its dimensionality. This results in a kernel matrix with low-dimensional structure lying in a large ambient space. A low-rank approximation, then, can capture all of the information in the kernel matrix without requiring $n^2$ entries.

\nystrom{} methods are one example of a general approach to computing low-rank matrix approximation from a subset of rows and/or columns of the matrix \cite{citeulike:12704915}. Choosing a relevant subset is broadly referred to as {\em column subset selection} (CSS). CSS methods have been applied successfully in applications ranging from image segmentation \cite{Fowlkes:2004:SGU:960255.960312} to genomic analysis \cite{pmid17151345} and matrix factorization \cite{MackeyTJ11}.

The success of CSS-based approaches for matrix approximation depends strongly on the columns chosen to approximate the range space of the matrix. Intuitively, uniform random sampling of the columns will provide an accurate approximation when the columns of the kernel matrix are independently and identically distributed. However, when the underlying data are non-uniformly distributed or even clustered, uniform sampling requires extra column draws to ensure an accurate approximation. In these settings, it has been shown in both theory \cite{deshpande2006matrix} and practice \cite{journals/jmlr/FarahatGK11} that adaptive sampling methods provide accurate approximations of low-rank kernel matrices with far fewer samples than uniform random sampling. In this way, we say that adaptive methods are more {\em efficient} than uniform random sampling.

Current adaptive sampling methods take advantage of the structure of the kernel matrix. Random adaptive sampling methods use the entries of the kernel matrix to compute a weighted probability distribution over the column indices. The distribution improves the chances of sampling relevant columns. Deterministic adaptive sampling methods use the already-sampled columns to compute a residual over the kernel matrix, from which a new column index is selected.

The downside of current adaptive methods is their computational burden. Adaptive methods do not scale well to large problem sizes for two reasons.  First, existing adaptive methods inspect the entire kernel matrix to determine which columns to select. For extremely large datasets, both forming and storing an explicit representation of the kernel matrix is intractable.   Second, existing adaptive methods require {\em dense} $n\times n$ matrix computations, even for {\em sparse} matrices that are otherwise easy to store because they have relatively few non-zeros elements. For these reasons, current adaptive methods cannot be applied to extremely large collections of data \cite{JMLR:v14:talwalkar13a}.

\subsection{Contributions}
For adaptive sampling methods, it is not generally possible to determine the best columns to select from a kernel matrix without explicitly forming all of the candidate columns. However, as the kernel matrix is symmetric, a small sample of columns provides partial information about the remaining un-sampled columns. Based upon this observation, we introduce a principled adaptive sampling scheme called Accelerated Sequential Incoherence Selection (oASIS) that predicts the most informative columns to sample from a kernel matrix without forming the entire matrix.  oASIS has several advantages over existing adaptive sampling schemes.  
\begin{itemize}
 \item oASIS does not require a fully precomputed kernel matrix. It can operate solely on the data, using the kernel function. This is because oASIS selects the column to be included in the approximation before explicitly computing it.  For this reason, only the submatrix of sampled columns must be computed/stored.   
 \item  oASIS's total runtime scales {\em linearly} with the matrix dimension $n,$ making it practical for large datasets.  For this reason, oASIS is orders of magnitude faster than other adaptive methods that have $\mathcal{O}(n^2)$ or higher runtime.
 \item oASIS can exactly recover the rank $r$ kernel matrix in $r$ steps.
 \item oASIS preserves zero entries in sampled columns of sparse kernel matrices, enabling greater efficiency in these cases.  This is in contrast to conventional greedy methods that require dense $n\times n$ matrix computations that ``fill in'' the zero entries in a matrix \cite{journals/jmlr/FarahatGK11}.  
\end{itemize}
oASIS provides a tractable solution for approximating extremely large kernel matrices where other adaptive methods simply cannot be applied \cite{SmoSch00, journals/jmlr/FarahatGK11, conf/icml/GittensM13}.  In a range of numerical experiments below, we demonstrate that oASIS provides accuracy comparable to the best existing adaptive methods \cite{SmoSch00, journals/jmlr/FarahatGK11, conf/icml/GittensM13, Zhang:2008:INL:1390156.1390311 } with dramatically reduced complexity and memory requirements.  

While oASIS is useful for kernel matrices, its usefulness becomes more pronounced when the dataset is so large that it can no longer be held entirely in memory. This is because we can parallelize oASIS by splitting up the dataset and the working matrices among various processors. We introduce an algorithm called oASIS-P that efficiently distributes the data and the submatrices used in approximation, as well as the computation and selection of a new column to add to the approximation. oASIS-P is highly scalable as it incurs minimum communication overhead in between the parallel computing nodes. We implemented oASIS-P using a standard message passing interface (MPI) \cite{gabriel04:open_mpi}. With oASIS-P, we can perform the \nystrom{} approximation in a data size regime where even the simplest algorithms become difficult to run. 

In addition, we study oASIS from a theoretical perspective and propose conditions under which oASIS will exactly recover a rank-$r$ kernel matrix in $r$ steps. Other greedy methods do not have this guarantee. oASIS can perform this recovery because it chooses linearly independent columns at each step, which enables efficient sampling. Random selection methods do not necessarily choose independent columns, and can choose redundant columns, resulting in inefficient sampling.

This paper is organized as follows. In Section~\ref{sec:bkgnd}, we introduce the \nystrom{} method, survey existing sampling methods, and describe important applications of kernel matrices in classification, clustering, and dimensionality reduction. In Section~\ref{sec:oASIS}, we describe the motivation behind our initial column sampling method, called Sequential Incoherence Selection or SIS. We then describe the accelerated version of SIS, or oASIS. We then describe a parallel version of oASIS, which we call oASIS-P. In Section~\ref{sec:theory}, we provide theory determining the conditions under which oASIS will exactly recover the kernel matrix. And in Section~\ref{sec:Experiments}, we use multiple synthetic and real datasets to demonstrate the efficacy of oASIS for approximating kernel matrices and diffusion distances for nonlinear dimensionality reduction \cite{Coifman20065}.

\section{Background}
\label{sec:bkgnd}
To set the stage, we will quickly describe a few common kernel and distance matrices used in machine learning. Following this, we introduce the \nystrom{} method and describe its variants and applications. 

\subsection{Notation}
We write matrices $G$ and vectors $x$ in upper and lowercase script, respectively. We use $A^\dagger$ to denote the Moore-Penrose pseudo-inverse of $A$. We represent the element-wise product of matrices $A$ and $B$ as $A\circ B.$  $\colsum(A)$ denotes a row vector, where the $i^{th}$ entry contains the sum of the $i^{th}$ column of $A.$ When describing algorithms, we use ``Matlab'' style indexing in which $G(i,j)$ denotes the $(i,j)$ entry of the matrix $G$ and $G(:,j)$ denotes its $j^{th}$ column. We use $\Lam$ to denote a collection of chosen indices of the columns of a matrix $A$; $\nLam$ is the set of indices not chosen. For example, $A_\Lam$ are all of the columns of $A$ indexed by the set $\Lam$.

\subsection{Kernel Matrices and their Applications}

Kernel methods are widely used in classification and clustering to ``lift'' datasets into a high-dimensional space where the classes of data are linearly separable.  This is done with the help of a {\em kernel function} $k(\cdot,\cdot)$ which measures pairwise similarities between points in the lifted space.  An $n\times n$ kernel matrix is then formed from $n$ data points $\{z_i\}_{i=1}^n$ with $G(i,j) = k(z_i,z_j),$ where high magnitude entries of $G$ correspond to pairs of similar data points.  The singular vectors of $G$ are then computed and used to map the data back into a low-dimensional space where the data is still linearly separable. The kernel trick is widely used in classification and clustering \cite{Williams01usingthe, Fowlkes:2004:SGU:960255.960312, Shi:2000:NCI:351581.351611, hastie_09_elements-of.statistical-learning, Simard:1992:EPR:645753.668226, Filippone:2008:SKS:1284917.1285173, KreBel:1999:PCS:299094.299108, Hastie:1998:CPC:302528.302744}

Manifold learning methods, including diffusion maps \cite{Coifman20065} and Laplacian eigenmaps \cite{LEM_NC_03}, map high-dimensional data that lie on nonlinear but low-dimensional manifolds onto linear low-dimensional spaces. 
These methods use a kernel matrix that encodes the {\em geodesic distance} between pairs of points --- the shortest path between two points along the surface of the underlying manifold. High-dimensional data are mapped into a low-dimensional space using the left singular vectors of $G$, and thus a singular value decomposition (SVD) of $G$ is required.  For a review of dimensionality reduction using geodesic and diffusion distance kernel matrices, see \cite{DBLP:journals/spm/TalmonCGC13, JMLR:v14:talwalkar13a}.

\subsection{The \nystrom{} Method} 
\label{sec:nystrom}

Williams and Seeger \cite{Williams01usingthe} first presented the \nystrom{} method  to improve the speed of kernel-based methods for classification. The method approximates a low-rank symmetric positive semidefinite (PSD) matrix using a subset of its columns. 
 
Consider an $n \times n$ PSD matrix $G$ of rank $r.$ For all PSD matrices $G$, there exists a matrix $X \in \mathbb R^{r \times n}$ such that $G = X^TX$. Suppose we choose a set $\Lam$ of $k \le r$ indices and then sample those $\Lam$ columns from $G$ as $C_k$, $C_k\in\mathbb{R}^{n \times k}$. We collect the $k$ indices into a set $\Lam$. The sampling forms a partition of $X=\begin{bmatrix} X_\Lam & X_\nLam \end{bmatrix}$.  We can then express $G$ as
\begin{equation} \label{eq:original}
G = \begin{bmatrix} X_\Lam^T \\ X_\nLam^T \end{bmatrix} \begin{bmatrix} X_\Lam & X_\nLam \end{bmatrix} = \begin{bmatrix}
W_k  & X_\Lam^TX_\nLam \\
X_\nLam^TX_\Lam & X_\nLam^TX_\nLam
\end{bmatrix},
\end{equation}
where $C_k = \begin{bmatrix} W_k  & X_\nLam^TX_\Lam \end{bmatrix}^T$ consists of the $n \times k$ sampled columns of $G$, and $W_k=X_\Lam^TX_\Lam$ is the $k \times k$ symmetric matrix containing the row entries of the sampled columns at the indices of the sampled columns. Note that without loss of generality we can permute the rows and columns of $G$ so that the columns in $C_k$ are the first $k$ columns of $G$. The \nystrom{} approximation of $G$ is defined as
\begin{equation}
\label{eq:definition}
G \approx \widetilde G_k = C_kW\dinv_kC_k^T.
\end{equation}
Note that neither $X$ nor any partition of $X$ is found explicitly, but that the $\widetilde{G}_k$ is found through the set of sampled columns $C_k$ and the respective rows $W_k$.

An approximate SVD of $G$ can be obtained from the SVD of $W_k,$ which is written as $W_k = U_W \Sigma_{W} U_W^T.$ The singular values ${\widetilde{\Sigma}}$ of the approximation $\widetilde G_k = \widetilde{U}\widetilde{\Sigma}\widetilde{U}^T$ are given by $(n/k)\Sigma_{W}$ \cite{Kumar:2012:SMN:2188385.2343678}, and the singular vectors are given by 
\begin{equation*}
\widetilde{U} = \sqrt{\frac{k}{n}}C_k U_{W}\Sigma_{W}^{-1}.
\end{equation*}
Since $W_k$ is $ k\times k,$ this computation is much faster than computing the full $n\times n$ SVD of $G$. The complexity of the SVD step reduces from $\mathcal{O}(n^3)$ to $\mathcal{O}(k^3)$ with $k \leq r \ll n$.

Note that the \nystrom{} method enables the singular vectors of $G,$ and thus a low dimensional embedding of the data, to be computed from only a subset of its columns.  When $n$ is large, it is desirable to form only a subset of the columns of $G$ rather than calculate and store all $n(n-1)/2$ pairwise kernel distances.

\subsection{Column Sampling Methods}
We now describe the four main categories of column selection methods. We compare oASIS with the methods listed below, as together they cover all of the types of sampling used currently in \nystrom{} approximation.

\subsubsection{Uniform Random Sampling}
Early work on the \nystrom{} method focused on random column sampling methods \cite{Williams01usingthe}. Theoretical bounds on the accuracy of a rank-$k$ approximation to $G$ after sampling a sufficient number of columns have been delveloped for various norms \cite{conf/icml/GittensM13}. 

Uniform random sampling is an appealing method as it is computationally fast. However, the accuracy of a matrix approximation depends directly on the specific columns sampled. This method does not take advantage of the structure inherent in the kernel matrix, leading to redundant sampling and larger approximation error. Improvements on this sampling method can be made by weighting the probability distribution used for the column draw, to increase the chance of selecting relevant columns.
 
\subsubsection{Non-deterministic Adaptive Sampling}
\label{sss:ndas}
Leverage scores are a recent method for computing the distribution over the column draw \cite{conf/icml/GittensM13}. Given the rank-$k$ SVD of $G_k = U_k\Sigma_kU_k^T$, the scores are computed as $s_j=\|U_k(j,:)\|^2$, and each column is selected with probability proportional to its score. This method provides accurate approximations by sampling more relevant columns. However, Leverage scores require the low-rank approximate SVD of $G$ to be precomputed at expensive $\mathcal{O}(n^3)$ cost. There are fast approximations available for finding the first few singular vectors and values of $G$ \cite{DBLP:journals/jmlr/DrineasMMW12}. Regardless, $G$ must be completely formed and stored before it is approximated. 

\subsubsection{Deterministic Adaptive Sampling}
\label{sss:dgm}
Early deterministic adaptive methods \cite{SmoSch00} use an exhaustive search to find columns that minimize the Frobenius norm error $\|G-\widetilde{G}_k\|_F$. While accurate, this method also requires a precomputed $G$, and has combinatorial complexity. A more efficient adaptive scheme by Farahat \cite{journals/jmlr/FarahatGK11} builds a \nystrom{} approximation by selecting columns sequentially using a matrix of ``residuals.'' At each stage of the method, the column with the largest residual is selected, and the residual matrix is updated to reflect its contribution.  While accurate, the method also requires a precomputed $G$, and the cost of updating the residual matrix is $\mathcal{O}(n^2)$ per iteration.

The residual criterion is related to the adaptive probability distribution calculated by Deshpande \cite{deshpande2006matrix}. After a sufficient number of columns are chosen, an orthogonalization step obtains a rank-$k$ approximation of $\widetilde{G}_k$.

\subsubsection{$K$-means \nystrom}
\label{sss:dkm}
Instead of approximating $G$ with direct column sampling, an approximation can be made from representative data points found with a $K$-means algorithm. A dataset consisting of clouds of points clustering around $K$ centroids can be described by finding the locations of the centroids. Each datapoint is then remapped into the eigenspace of the centroids. This method was first described by Zhang \cite{Zhang:2008:INL:1390156.1390311}. Since the computed centroids do not exist in the dataset, the method does not directly sample columns, but remaps them onto a rank-$K$ space. Once the solution to the $K$-means is found, the remapping is $\mathcal{O}(\ell n)$. While finding an exact solution to $K$-means is {NP-hard}, generally $K$-means will converge in $\mathcal{O}(n^2)$ time. The resulting $\wtg$ can not be formed from the columns of $G$, and so has no space saving representation. 

In a survey of methods by Kumar \cite{Kumar:2012:SMN:2188385.2343678}, $K$-means was found to be the state-of-the-art approximation method compared to previous sampling methods such as Incomplete Cholesky Decomposition \cite{Fine01efficientsvm}, Sparse Matrix Greedy Approximation \cite{SmoSch00}, and Kumar's Adaptive Partial method derived from Deshpande's Adaptive Sampling method \cite{deshpande2006matrix}. For this reason, in lieu of comparisons with many different adaptive sampling techniques we can compare our results directly with $K$-means. For our experiments, we used the same code as provided in \cite{Zhang:2008:INL:1390156.1390311}, with parameters used in both \cite{Zhang:2008:INL:1390156.1390311} and \cite{Kumar:2012:SMN:2188385.2343678}.

\subsection{Finding Low-Dimensional Structure}
\label{ss:otherapps}
In addition to using CSS for low-rank kernel approximation, column selection approaches have also been used to find important data points and in turn, reveal low-dimensional structure in data \cite{Mahoney20012009, journals/siamsc/ChengGMR05}. Recently, it was shown in \cite{Dyer2015SEED} that oASIS can be used to select representative examples from large datasets in order to enable clustering and dimensionality reduction. This method, called Sparse Self-Expressive Decomposition (SEED), consists of two main steps. First, oASIS is used to select data points for a dictionary. Second, all of the data is then represented by a sparse combination of the points in the dictionary using Orthogonal Matching Pursuit \cite{davisOMP,rubinstein2008efficient}. The sparsity patterns found in the representations can be used for for clustering, denoising, or classification. SEED's ability to properly describe data hinges on the selection of data points used for the dictionary. oASIS is able to efficiently sample good points to use for this task, compared to other adaptive sampling methods. A full treatment of SEED is can be found in \cite{Dyer2015SEED}.
 
\section{Accelerated Sequential Incoherence Selection (oASIS)}
\label{sec:oASIS}
In this section, we introduce oASIS, a new adaptive sampling method for column subset selection. We also introduce a parallel version called oASIS-P, and we analyze the complexity of both.

\subsection{Sequential Incoherence Selection (SIS)} \label{section:ma}
We now address the question of how to build a \nystrom{} approximation by sequentially choosing columns from $G.$
Suppose we have already selected $k$ columns from $G$ and computed a \nystrom{} approximation $\widetilde{G}_k.$ Our goal is to select a new column that improves the approximation.  If a candidate column lies in the span of the columns that we have already sampled, then adding this column has no effect on $\|G-\wtg_k\|$.  Ideally, we would like to quantify how much each candidate column will change the existing approximation and select the column that will produce the most significant impact. Since an ideal column does not lie in the span of the columns that have been selected, we say that this column should be {\em incoherent} with those already selected.

We can develop a criteria for finding this new, incoherent column of $G$ as follows. Consider a PSD matrix $G$. Recall that any such $G$ can be written as $X^TX$, where $X$ contains $n$ points in $r$ dimensions. Given an index set $\Lam$ of $k$ columns we can form a partition $X=\begin{bmatrix} X_\Lam & X_\nLam \end{bmatrix}$, and can collect the $k$ selected columns from $G$ into a matrix ${C_k = G_\Lam = \begin{bmatrix} W_k  & X_\nLam^TX_\Lam \end{bmatrix}^T}$. To improve the approximation $\wtg_k$, we select the best new column from $G$, append it to $C_k$, and compute a new $\wtg_{k+1}$. The best new column index $i\in\nLam$ in $G$ directly corresponds to the index of the column $x_i$ that lies farthest from the span of $X_\Lam$. This column $x_i$ satisfies
\eqb{bestinspan2}
\argmax_{i\in\nLam}{\|(I-P_\Lam)x_i\|_2^2},
\eqe
where $I$ is the identity matrix and $P_\Lam=X_\Lam X_\Lam\dinv$ is an orthogonal projection onto the span of the $k$ columns in $X$ that have already been selected. Provided that the columns in $X_\Lam$ are linearly independent, we can expand \eqref{bestinspan2} as
\eqb{bestinspan3}
\argmax_{i\in\nLam}{x_i^Tx_i - x_i^TX_\Lam(X_\Lam^TX_\Lam)\inv X_\Lam^Tx_i}.
\eqe

\begin{figure}[!t]
\begin{center}
\includegraphics[scale=.4,  trim = 0mm 0mm 0mm 0mm, clip]{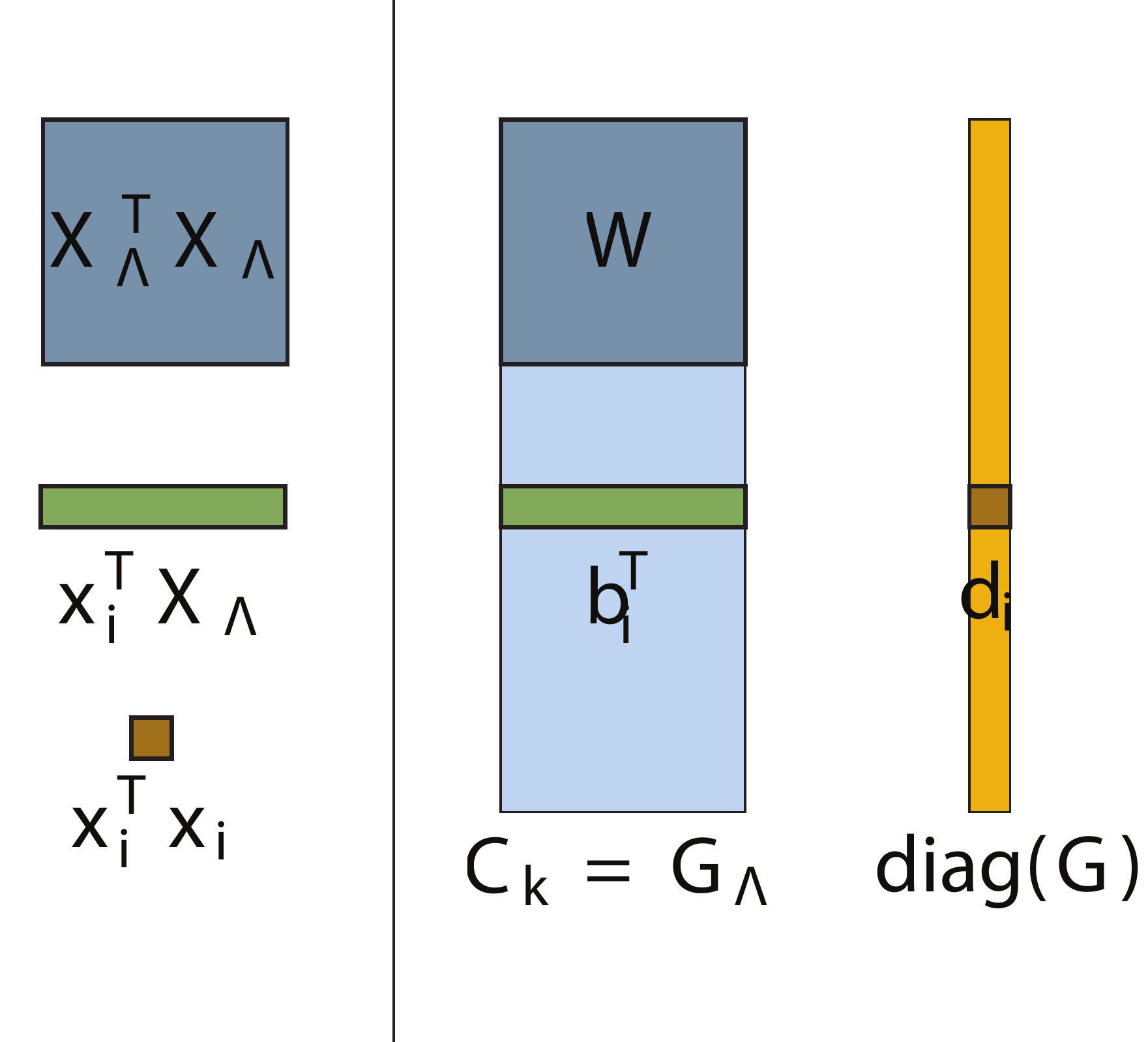} 
\end{center}
\caption{All of the terms in the criterion in \eqref{bestinspan2} for the next column index to select can be found in the structure of the columns $C_k=G_\Lam$ that have already been sampled, and the diagonal of $G$. This allows for a deterministic sampling without precomputing all of $G$. See Section~\ref{section:ma} for details.}
\label{fig:factorization}
\end{figure}

Even though $X$ is not known explicitly, \eqref{bestinspan3} can be evaluated based upon knowledge of $C_k$ and $\diag(G)$. The first term of the expression in \eqref{bestinspan3} is the diagonal entry $i$ of $G$, which we denote as $d_i$. The second term can be written as $b_i^TW_k^{\inv}b_i$, where $b_i^T = x_i^TX_\Lam$ is one of the $n-k$ rows of $C_k$ indexed by $\nLam$, and $W_k$ is comprised of the $k$ rows of $C_k$ indexed by $\Lam$. When $X_\Lam$ contains linearly dependent columns, we replace $W_k\inv$ with $W_k\dinv$. Therefore, we can iteratively select columns to sample for the approximation {\em without} precomputing the entire kernel matrix $G$, as shown in Figure~\ref{fig:factorization}. This sets oASIS apart from all other adaptive methods, as discussed in Sections~\ref{sss:ndas} and~\ref{sss:dkm}.

With the evaluation of our criteria now possible, we develop the following sampling method for sequential incoherent selection (SIS).  We assume that the process has been seeded with a small set $\Lam$ of $k_0$ column indices chosen at random.  Columns are then selected to be included in the approximation as follows:  
\begin{enumerate}
\item Let $k=|\Lam|$. Collect the columns of $G$ indexed by $\Lam$ as $C_k$. Form  $W_k\inv$ from the rows of $C_k$ indexed by $\Lam$. 
 
\item Let $b_i^T$ denote row $i$ of $C_k$, and let $d_i$ denote element $i$ of $\diag(G)$. For each unselected column $i \in \nLam$, calculate
   $$\Delta_i =  d_i - b_i^TW_k\inv b_i  .$$
\item Select the column that maximizes $| \Delta_i|$ and set ${\Lam \leftarrow \Lam \cup \{i\}}$.  
\item If the selected value of $|\Delta_i|$ is smaller than a user set threshold, then terminate.  Otherwise, return to Step 1.
 \end{enumerate}

\subsection{oASIS}

A naive implementation of SIS in Section \ref{section:ma} is inefficient, because each step requires a matrix inversion to form $W_k\inv$ in addition to calculating the errors $\Delta_i.$  Fortunately, both of these calculations can be performed efficiently by updating the results from the previous step using block matrix inversion formulas. We dub this new method oASIS.
  
We first consider the calculation of $W\kp\inv$ after a new column is added to the approximation made from $k$ columns.   We assume throughout the rest of this section that $W\kp$ is invertible and thus $W\kp^\dagger = W\kp\inv.$ We show that our column selection rule guarantees the invertibility of $W_k$ in Section~\ref{sec:guarantees}. 
 Let $b$ denote the first $k$ entries of the new column, $d$ denote the relevant element of $\diag(G)$, and $\Delta\kp =d-b^TW_k\inv b$.  Using a block inversion formula, we obtain
 \begin{align}\label{update1}
  W\kp\inv = 
 \begin{bmatrix}
W_k & b \\
b^T  &d
\end{bmatrix}\inv  
=& 
\begin{bmatrix}
 W\inv_k+s q q^T 	& 	-s q  \\
 -s q^T		& 	s
 \end{bmatrix},
\end{align}
where $s = (d - b^T W\inv_k b)\inv = \Delta\kp\inv $ is the (scalar valued) Schur complement and $q=W\inv_k b$ is a column vector.  This update formula enables $W\kp\inv$ to be formed by updating $W_k\inv$  and only requires inexpensive vector-vector multiplication.  Note that $W\kp$ is invertible as long as $\Delta\kp$ is non-zero, which is guaranteed since the algorithm terminates if $\Delta\kp = 0$, in which case our approximation is exact.  

Now consider the calculation of $\Delta_i=d_i - b_i^TW_k^\dagger b_i $ for all candidate columns $i.$    Note that $C_k^T=[b_1, b_2,\cdots,b_n].$ We can evaluate all $b_i^TW_k^\dagger b_i $ simultaneously by computing the entry-wise product of $C_k$ with the matrix $R_k:=W\inv_k C_k^T$ and then summing the resulting columns. If we have already formed $C_k$ and $R_k$, then the matrix $R\kp = W\inv\kp C\kp^T$ needed on the next iteration is obtained by applying \eqref{update1} to $C\kp^T$ to obtain    
\begin{align}\label{update2}
 R\kp &= W\inv\kp C^T\kp=W\inv\kp 
 \begin{bmatrix}
C^T_k \\
c\kp^T
 \end{bmatrix} \\ \nonumber
&= 
 \begin{bmatrix}
R_k+ s q  (q^T C^T_k-c^T\kp) \\
s(-q^T C^T_k+c^T\kp)
 \end{bmatrix}.
\end{align}

Equation \eqref{update2} forms $R\kp$ by updating the matrix $R_k$ from the previous iteration.  The update requires only matrix-vector and vector-vector products.   The application of this fast update rule to the method described in Section \ref{section:ma} yields oASIS, detailed in Figure~\ref{alg:oasis}.  

 oASIS can be initialized with a small random subset of $k_0$ starting columns from $G.$  Next, the starting matrices $C_k,$ $W\inv_k$ and $R_k = W\inv_kC^T_k$ are formed.  On each iteration of the algorithm, the vector of Schur complements $\Delta$ is formed by computing
   $$\Delta = d - \colsum(C_k \circ R_k).$$
Next, the largest entry in $\Delta$ is found, and its index is used to select the next column from $G.$  The update formulas \eqref{update1} and \eqref{update2} are then used to form the matrices $W\inv\kp$ and $R\kp$ required for the next iteration.
 
\begin{figure}[!t]
   \begin{algorithmic}
  	\STATE{\bfseries Algorithm 1: oASIS}
	\STATE{\bfseries Inputs:} Symmetric matrix $G \in\mathbb{R}^{n \times n}$,
	\STATE \hspace{1.1cm} Diagonal elements of $G$, stored in $d$, 
	\STATE \hspace{1.1cm} Maximum number of sampled columns, $\ell$, 
	\STATE \hspace{1.1cm} Initial number of sampled columns, $k<\ell,$ 
	\STATE \hspace{1.1cm} Non-negative stopping tolerance, $\epsilon.$ 
	\STATE{\bfseries Outputs:} The sampled columns $C,$
	\STATE \hspace{1.37cm} The inverse of the sampled rows $W\inv.$
	\STATE{\bfseries Initialize: } Choose a vector $\Lam \in [1,n]^\ell$ of $k$ random starting indices. 
	\STATE $C_k = G(:, \Lam)$ 
	\STATE $W\inv_k = G(\Lam, \Lam)\inv$
	\STATE  $R_k = W\inv_kC^T_k$
	\WHILE{$k<\ell$} 
	\STATE $\Delta = d - \colsum(C_k \circ R_k)$
	\STATE $i = \argmax_{j \not\in \Lam}  |\Delta(j)|$
	\IF{$|\Delta(i)|<\epsilon$}
	\STATE {\bfseries return}
	\ENDIF 
	\STATE $b  = G(\Lam,i)$ 
	\STATE $d = d(i)$
	\STATE $s = 1/\Delta(i)  $
	\STATE $q = R(:,i)$
	\STATE $C\kp = [C_k,  G(:,i)]$
	\STATE Form $W\inv\kp$ using \eqref{update1}
	\STATE Update $R\kp$ using \eqref{update2}
	\STATE $k\leftarrow k+1$
 	\STATE $\Lam \leftarrow \Lam \cup \{i\}$
	\ENDWHILE
	\end{algorithmic}
   \caption{The oASIS algorithm. Note that $G$ need not be precomputed.}
   \label{alg:oasis}
\end{figure}

\subsection{Parallel oASIS}
\label{ss:parallizedoasis}
\begin{figure}[!t]
\begin{center}
\includegraphics[scale=.4,  trim = 0mm 0mm 0mm 0mm, clip]{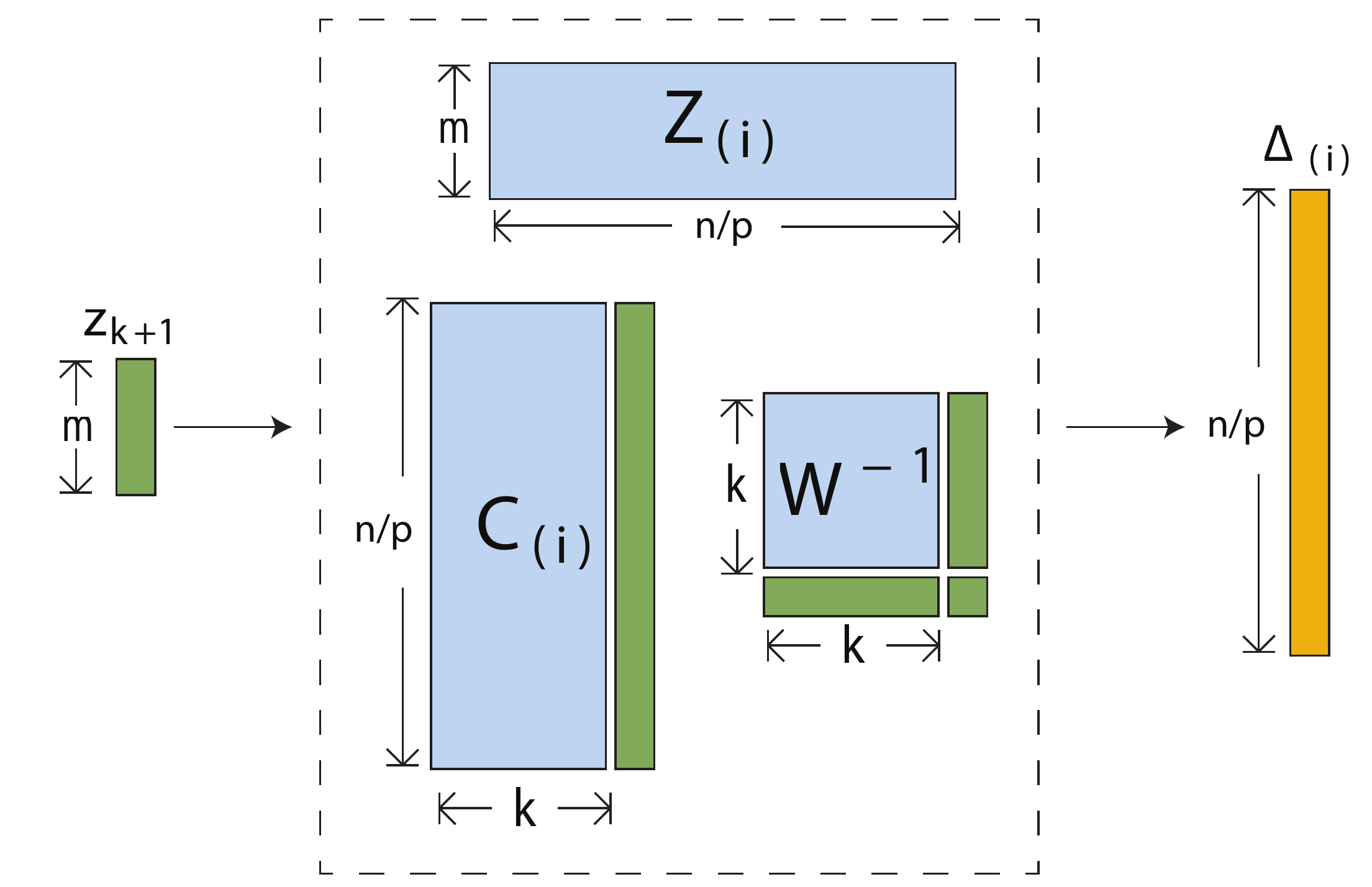} 
\end{center}
\caption{Diagram of a single node in oASIS-P. Each node retains a subset $Z_{(i)}$ of the data. The node receives the selected data point $z_{k+1}$, and updates $C_{(i)}$ and $W\inv$ using the kernel function $g(z_i,z_j)$ as in Fig.~\ref{alg:oasispar}. It then computes and broadcasts $\Delta_{(i)}$ to determine the next point to sample from the entire dataset. See Section~\ref{ss:parallizedoasis} for details.}
\label{fig:oASISPdiagram}
\end{figure}

For the case of kernel matrices $G$ generated from a dataset $\{z_i\}_{i=1}^n$, oASIS does not need to explicitly store $G$. Therefore, oASIS saves time, as computing the full $G$ is expensive. oASIS also saves space, as the full $G$ increases quadratically with $n$. However, the dataset may be itself very difficult to store due to size. In addition, oASIS requires space to build $C$, $W\inv$, and $R$. In total, oASIS requires $\mathcal{O}(mn+\ell^2+2\ell n)$ of memory. In cases where dataset is too large to fit in memory, the matrix operations for oASIS can be distributed among $p$ separate processor nodes. We begin by arranging the dataset columnwise into a matrix $Z$. Each node stores a submatrix $Z_{(i)}$ consisting of $n/p$ columns of $Z$, a copy of $W\inv$, and the column entries of $C_k$ and $R_k$ corresponding to the results of the kernel function over the entries in $Z_{(i)}$ with the entries in $Z_{\Lam}$. When a new column index $i$ is selected, the node storing column vector $z_i$ is found and the column is broadcast to all of the nodes. Each node can then calculate the appropriate new entries as needed, as shown in Fig.~\ref{fig:oASISPdiagram}. For each new column, the size of the communicated vector is the dimensionality of the data point, which is much smaller than the kernel matrix. Low communication overhead is an essential property of oASIS-P since, in distributed settings, the cost of internode communication can be larger than intranode computation. The memory requirements for over each node becomes $\mathcal{O}(mn/p+\ell^2+2\ell n/p + \ell m)$, which makes performing oASIS over datasets with millions of points tractable.

We call this method Parallel oASIS, or oASIS-P, and it is detailed in Figure~\ref{alg:oasispar}. The implementation makes use of the standard MPI commands $Broadcast(data)$ (to send data from one node to every node) and $Gather(variable)$ (to concatenate variables in each node into a single variable on one central node).

\begin{figure}[!t]
	\begin{algorithmic}
	\STATE{\bfseries Algorithm 2: oASIS-P}
	\STATE{\bfseries Inputs:} Dataset arranged as a matrix $Z\in\mathbb{R}^{m \times n}$
	\STATE \hspace{1.1cm} Kernel function $g(z_i,z_j)$,
	\STATE \hspace{1.1cm} Number of nodes $p$ indexed by $(i)$,
	\STATE \hspace{1.1cm} Maximum number of sampled columns, $\ell,$ 
	\STATE \hspace{1.1cm} Initial number of sampled columns, $k<\ell,$ 
	\STATE \hspace{1.1cm} Number $n$ of columns in $Z$,  
	\STATE \hspace{1.1cm} Non-negative stopping tolerance, $\epsilon.$ 
	\STATE{\bfseries Outputs:} The sampled columns $C,$
	\STATE \hspace{1.37cm} The inverse of the sampled rows $W\inv.$
	\STATE{\bfseries Initialize: } Choose a vector $\Lam \in [1,n]^\ell$ of $k$ random 
	\STATE \hspace{1.6cm} starting indices. 
	\STATE \hspace{1.6cm}  Load separate $n/p$ column blocks of $Z$ into
	\STATE \hspace{1.6cm} each node as $Z_{(i)}$.
	\STATE \hspace{1.6cm}  $Broadcast(Z(:,\Lam))$ as $Z_\Lam$.	

	\STATE {\bfseries On each node $(i):$}
	
	\STATE $d_{(i)}(j)$ = ${g(Z_{(i)}(:,j),Z_{(i)}(:,j))}$ over all $j$ in $Z_{(i)}$
	\STATE $C_{(i)}(i,j)$ = ${g(Z_{(i)}(:,i),Z_\Lam(:,j))}$ over all $i$ in $Z_{(i)}$ 
	\STATE \hspace{0.3cm} and all $j$ in $Z_\Lam$
	
	\STATE $W(i,j)$ = ${g(Z_\Lam(:,i),Z_\Lam(:,j))}$ over all $i$ in $Z_\Lam$ 
	\STATE \hspace{0.3cm} and all $j$ in $Z_\Lam$
	\STATE Compute $W\inv$ 	

	\STATE  $R_{(i)} = W\inv C^T_{(i)}$
	
	\WHILE{$k<\ell$} 
	\STATE {\bfseries On each node $(i):$}
	
	\STATE $\Delta_{(i)} = d_{(i)} - \colsum(C_{(i)} \circ R_{(i)})$
	
	\STATE  $\Delta = Gather(\Delta_{(i)})$
	\STATE $i = \argmax_{j \not\in \Lam}  |\Delta(j)|$
	
	\IF{$|\Delta|<\epsilon$}
	\STATE {\bfseries return}
	\ENDIF 
	\STATE $z_{k+1} = Broadcast(Z(:,i))$
	\STATE $\Delta = Broadcast(\Delta)$
	\STATE {\bfseries On each node $(i):$}

	\STATE $c_{(i)k+1}$ = ${g(Z_{(i)}(:,i),z_{k+1})}$ over all $i$ in $Z_{(i)}$ 
	\STATE $C_{(i)}= [C_{(i)},  c_{(i)k+1}]$

	\STATE $q = g(Z_\Lam(:,i),z_{k+1})$ over all $j$ in $Z_\Lam$

	\STATE $s = 1/\Delta $

	\STATE Update $W\inv$ using \eqref{update1}
	\STATE Update $R_{(i)}$ using \eqref{update2}
	
	\STATE $Z_{\Lam}= [Z_\Lam,  z_{k+1}]$

 	\STATE $k\leftarrow k+1$
 	\STATE $\Lam \leftarrow \Lam \cup \{i\}$
	\ENDWHILE
	
	\STATE $W\inv = W\inv_{(i)}$
	\STATE $C = Gather(C_{(i)})$

	\end{algorithmic}
   	\caption{oASIS-P, for datasets too large for a single node.}
   	\label{alg:oasispar}
\end{figure}

\section{Properties and Applications of oASIS} 
\label{sec:theory}

\subsection{Theoretical Guarantees of oASIS}
\label{sec:guarantees}
For general PSD matrices $G$ of rank $r$, we can guarantee that oASIS will finish in $r$ steps. We develop the theory needed to prove this guarantee in Sections~\ref{sec:isp} and \ref{sec:emr}. When $G$ is a Gram matrix, formed as $G(i,j) = {z_i}^Tz_j$, we can use the sample index set $\Lambda$ found with oASIS to make additional guarantees on approximating the dataset itself. We mention the guarantees briefly in Section~\ref{sss:oasisgram}, and they are described fully in \cite{Dyer2015SEED}. 

In Section~\ref{sec:isp}, we show that oASIS will select linearly independent columns of $G$ at each step. This becomes very useful in practice, as $\wtg$ is computed from the $W\dinv$, where $W=X_\Lam^TX_\Lam$. If the selected columns of $G$ are not independent, then $W$ is a singular matrix. By selecting linearly independent columns of $G$, oASIS can guarantee that $W\dinv = W\inv$, enabling time and space saving calculation of this element when computing $\wtg$. We discuss this further in Section~\ref{sss:othertheory}.

\subsubsection{Independent Selection Property of oASIS}
\label{sec:isp}
Given a PSD matrix $G = X^T X$,  ${\rm rank}(G) = {\rm rank}(X) = r$. In Lemma~\ref{thm:indep} below, we provide a sufficient condition that describes when oASIS will return a set of $r$ linearly independent columns. This is a similar condition as that provided in \cite{Dyer2015SEED}, although we provide an alternate proof.

\begin{LEMM} \label{thm:indep} At each step of Alg. \ref{alg:oasis}, the $i^{th}$ column of the matrix $G$ is linearly independent from the previously selected columns provided that $\Delta(i) > 0$.
\end{LEMM}

\begin{IEEEproof}
We prove this by construction of $X_\Lam$. Consider adding a new column $x_i$ to $X_\Lam$ with nonzero $\Delta(i) = {\|(I-P_\Lam)x_i\|_2^2}$ and $i\in\nLam$. Then $x_i$ is linearly independent of each column in $X_\Lam$, and so $G_{k+1}=X^Tx_i$ is linearly independent from any other $G_{j}=X^Tx_j$. Therefore as long as $\Delta(i) > 0$ at each step, the column selected will be linearly independent from the previous columns selected. 
\end{IEEEproof}

{\bf Remark.} This result guarantees that oASIS will return a set of $r$ linearly independent columns in $r$ steps as long as the selection criterion $\Delta(i) \ne 0$ holds before exact reconstruction occurs. While the algorithm may terminate early if $\Delta(i) = 0$ before $r$ columns have been selected, we have not observed this early termination in practice. 

\subsubsection{Exact Matrix Recovery}
\label{sec:emr}
We now prove that when oASIS selects $r$ columns from $G$, then $\wtg = G$.

\begin{THEO} \label{thm:exact} If oASIS chooses $r$ columns from a PSD matrix $G$ with ${\rm rank}(G)=r$, the \nystrom{} $\wtg = G$.
\end{THEO}
\begin{IEEEproof} As $X$ is rank $r$, and oASIS has chosen $r$ linearly independent columns, then at the next step all $\Delta(i)=0$ as ${\|(I-P_\Lam)x_i\|_2^2} = 0 \forall {i}$. Therefore ${\|(I-X_\Lam X_\Lam{\dinv})x_i\|_2^2} = 0$, or ${\|X-X_\Lam X_\Lam{\dinv}X\|_F} = 0.$ $X=\begin{bmatrix} X_\Lam & X_\nLam \end{bmatrix}$, and therefore ${\|X_\nLam-X_\Lam X_\Lam{\dinv}X_\nLam\|_F} = 0$, or $X_\nLam = X_\Lam X_\Lam{\dinv}X_\nLam.$ 
Expanding $\wtg$ in terms of $X_\Lam$ and $X_\nLam$,
\begin{align} \nonumber
\wtg = CW\dinv C^T 
&=  
\begin{bmatrix} X_\Lam^TX_\Lam \\  X_\nLam^TX_\Lam \end{bmatrix} 
\begin{bmatrix} X_\Lam^TX_\Lam\end{bmatrix} \dinv 
\begin{bmatrix} X_\Lam^TX_\Lam & X_\Lam^T X_\nLam \end{bmatrix} \\ \nonumber
&=
\begin{bmatrix} X_\Lam^TX_\Lam  & X_\Lam^T X_\nLam \\
X_\nLam^TX_\Lam & X_\nLam^TX_\Lam (X_\Lam^TX_\Lam)\dinv X_\Lam^T X_\nLam
\end{bmatrix}.
\end{align}
We examine the lower right block of this expansion as the others exactly match that of $G$ in \eqref{eq:original}. 

By Lemma~\ref{thm:indep} $X_\Lam$ is full rank, and so 
\begin{align} \nonumber
X_\nLam^TX_\Lam (X_\Lam^TX_\Lam)\dinv X_\Lam^T X_\nLam 
&=  
X_\nLam^TX_\Lam (X_\Lam^TX_\Lam)\inv X_\Lam^T X_\nLam \\ \nonumber
&=  
X_\nLam^T(X_\Lam X_\Lam\dinv) X_\nLam  \\ \nonumber
&= X_\nLam^T X_\nLam.
\end{align}
Thus the expansion of $\wtg$ is equal to the expansion of $G$ in \eqref{eq:original}. 
\end{IEEEproof}

\begin{figure*}[t]
\begin{center}
\subfloat[ ]{%
\includegraphics[scale=0.5, trim = 0mm 0mm 0mm 0mm, clip]{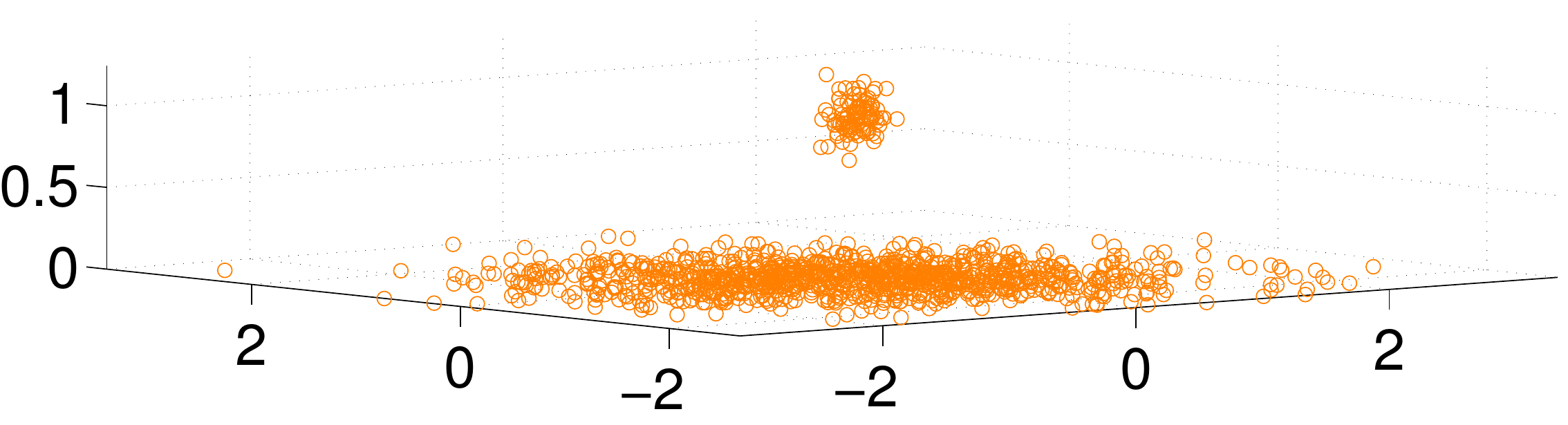}}\quad~
\subfloat[ ]{%
\includegraphics[scale=0.4, trim = 0mm 0mm 0mm 0mm, clip]{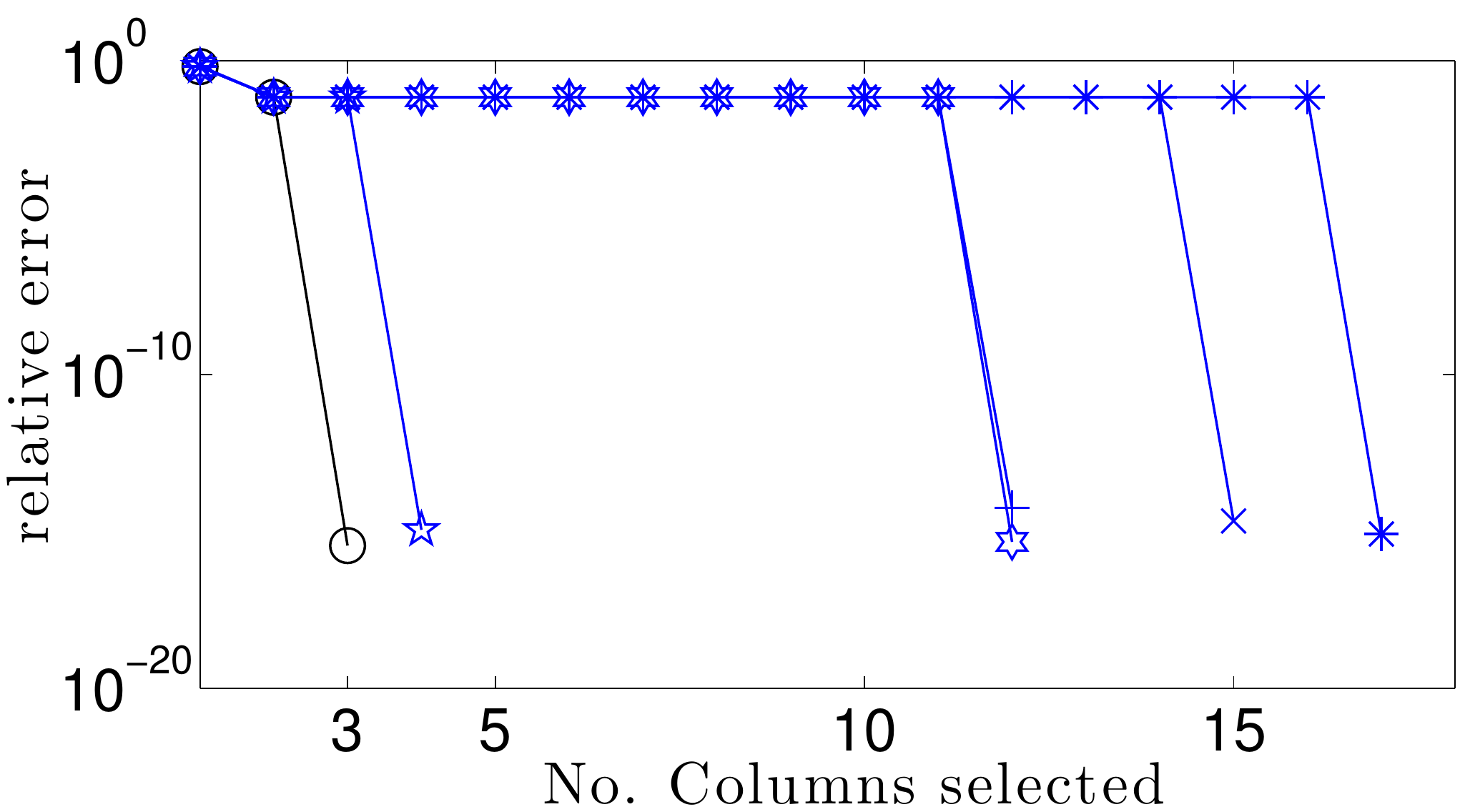}}
\subfloat[ ]{%
\includegraphics[scale=0.38, trim = 0mm 0mm 0mm 0mm, clip]{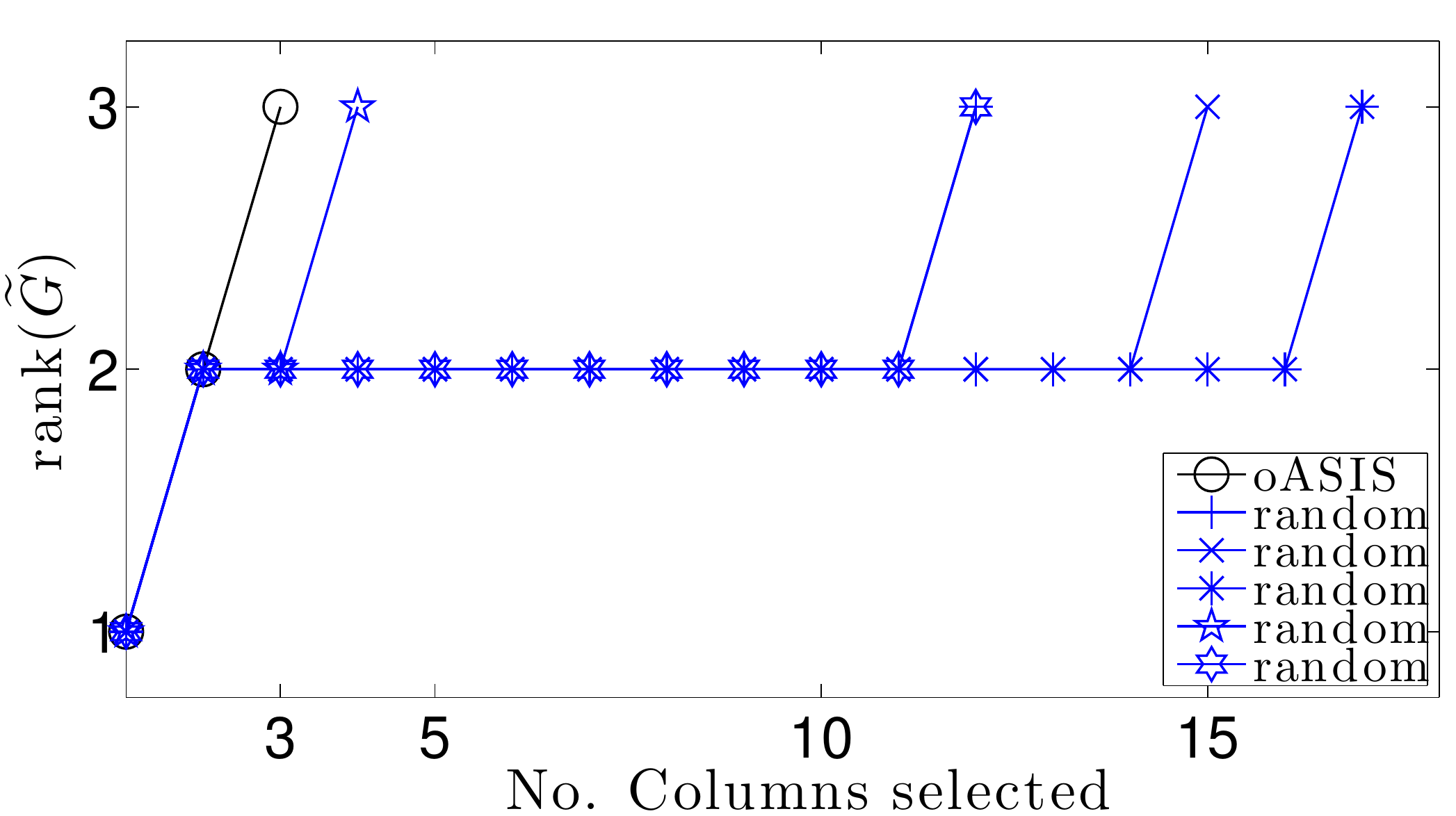}}
\end{center}
\caption{For a PSD Gram matrix $G$ formed from the dataset in (a), we compare approximation errors for $\wtg$ formed using oASIS vs. 5 separate trials of uniform random sampling in (b). We terminate trials at exact recovery. oASIS guarantees exact recovery after sampling 3 columns. Random sampling chooses redundant columns, and this inefficient sampling results in less accurate approximations. For uniform random sampling many of the error curves lie directly on top of each other, as the method repeatedly chooses redundant columns from the bottom cluster of data. The variety of columns sampled can be illustrated by plotting the rank of $\wtg$ by number of sampled columns in (c). oASIS chooses columns that increases the rank of $\wtg$ at each step. In contrast, choosing redundant columns results in a rank-deficient approximation. See Section~\ref{sss:othertheory} for details.}
\label{fig:exactrecovery}
\end{figure*}

\subsubsection{Guarantees for oASIS when $G$ is a Gram Matrix}
\label{sss:oasisgram}
When $G$ is a Gram matrix, we can arrange the points in the dataset $\{z_i\}_{i=1}^n, z_i \in \mathbb{R}^m$ columnwise into a matrix $Z$. Then we can write $G=Z^TZ$, with $Z$ of rank $m$. oASIS selects a set $|\Lambda|=m$ such that $Z=P_\Lambda(Z)$ exactly. This property is useful in solving a more general CSS problem than \nystrom, precisely formulated as
\begin{equation} \label{eq:cssgeneral}
\min_{|\Lambda| = L}\|  Z - P_{\Lambda}{Z}  \|_F,
\end{equation}
where $Z$ is an $m \times n $ matrix. This problem has combinatorial complexity, and many of the selection schemes for \nystrom{} arise from attempting to solve this more general problem. Indeed, the adaptive selection methods in \cite{journals/jmlr/DrineasM05, deshpande2006matrix}, and others, can also apply to this problem. We have developed guarantees on oASIS's ability to exactly recover $Z$ so that we can use the columns in $Z_\Lambda$ in developing a self-expressive decomposition of $Z$. Although a full treatment of SEED is described in \cite{Dyer2015SEED}, we briefly describe this extension in Section~\ref{ss:otherapps}.

\subsubsection{Comparison with Other Theory}
\label{sss:othertheory}
oASIS can guarantee exact matrix recovery in an information theoretically efficient number of steps. We show a synthetic example in Figure~\ref{fig:exactrecovery}. Using a dataset consisting of points drawn from a $2D$ Gaussian distribution centered on $(0,0)$ and points drawn from a $3D$ Gaussian distribution centered on $(0,0,1)$, we compute a PSD Gram matrix $G$ with a resulting rank$(G)=3$. oASIS selects columns linearly independent from previously selected columns at each step, increasing the rank of the approximation each time. This enables oASIS to use an iterative update of $W\inv$ instead of computing $W\dinv$ after all columns have been selected. At 3 steps, oASIS terminates, with $\wtg=G$ within machine tolerance.

Random or adaptive random sampling techniques have theoretical guarantees that $\wtg$ will be close to a rank-$k$ approximation of $G$ after a certain number of iterations \cite{conf/icml/GittensM13, journals/jmlr/DrineasM05}. However, the lack of guarantees on column selection make for redundant sampling. As an illustration, we include separate trials of uniform random sampling in Figure~\ref{fig:exactrecovery}. Uniform random sampling frequently selects columns within the span of previously selected columns at each step, and as a result the approximation error is generally higher than oASIS. In cases of very large data, this becomes a practical concern in computing both $C$ and $W\dinv$.

\subsection{Complexity of oASIS}
\label{sec:advantage}
The rate-limiting step of oASIS in Fig.~\ref{alg:oasis} is the computation of $R\kp$ by updating $R_k.$ Equation \eqref{update2} enables this to be performed by sweeping over the entries of $R_k,$ which has dimension $k\times n.$   The complexity of a single iteration is thus $\mathcal{O}(kn).$  If $\ell$ columns are sampled in total, then $\sum_{k=1}^\ell kn=\frac{1}{2} \ell(\ell+1)n$ entries must be updated.  The resulting complexity of the entire oASIS algorithm is thus $\mathcal{O}(\ell^2n).$ In practice, the number of sampled columns $\ell$ is much less than $n.$  This makes oASIS considerably more efficient than adaptive methods such as Farahat's \cite{journals/jmlr/FarahatGK11}, which requires the computation of $n\times n$ residual matrices at each stage resulting in $\mathcal{O}(\ell n^2)$ complexity.  oASIS is also more efficient than Leverage scores \cite{conf/icml/GittensM13}, since the scores use an approximate SVD of $G$ that requires $\mathcal{O}(n^2)$ computations over dense matrices. oASIS is about as efficient as $K$-means \nystrom, with complexity $\mathcal{O}(\ell n)$. However, $K$-means does not select columns, and instead forms the full $\wtg$ from the low-dimensional remapping. As a result, while $K$-means is useful in \nystrom{} approximation, it may not be as useful for more general CSS methods. oASIS, in contrast, can be used in more general CSS problems via the Gram matrix. 

If we only compare the speed in finding $\Lam$, oASIS is much slower than uniform random sampling, with its $\mathcal{O}(1)$ sampling speed. But oASIS also computes $C$ and $W\inv$ along the way, while these still need to be computed after selecting the columns to be used under uniform random sampling. In large data regimes, these become practical considerations that make implementation of uniform random sampling less efficient, as we discuss in Section~\ref{ss:cmplxoasisp}.

\subsection{Complexity of oASIS-P} 
\label{ss:cmplxoasisp}
The low complexity of oASIS makes it practical for extremely large matrices where other adaptive sampling schemes are intractable. For oASIS-P, the computational complexity of oASIS is divided by the $p$ nodes, such that each node has $\mathcal{O}(\ell^2 n/p)$ complexity. At first blush, oASIS-P is still slower than uniform random sampling and its $\mathcal{O}(1)$ sampling speed. However, in regimes where the dataset cannot be loaded entirely in memory, three practical considerations make uniform random sampling less competitive. First, forming $\ell$ columns from a dataset $\{z_i\}_{i=1}^n$ with $z_i \in \mathbb{R}^m$ takes at least $\mathcal{O}(\ell n)$ time. For many applications, forming the columns as they are sampled is substantially more expensive than the process of adaptively selecting columns. Second, communication of data vectors among nodes becomes the bottleneck in the parallel implementation, and oASIS-P and random sampling appear competitive in terms of column selection/generation time, as shown in Table~\ref{tab:ptable}.  Third, random sampling may require substantially more columns than oASIS to achieve the same accuracy, in which case adaptive sampling is highly advantageous.

\section{Numerical Experiments}
\label{sec:Experiments}
\begin{figure*}[!t]
\begin{center}
\subfloat{%
\includegraphics[scale=0.4, trim = 0mm 0mm 0mm 0mm, clip]{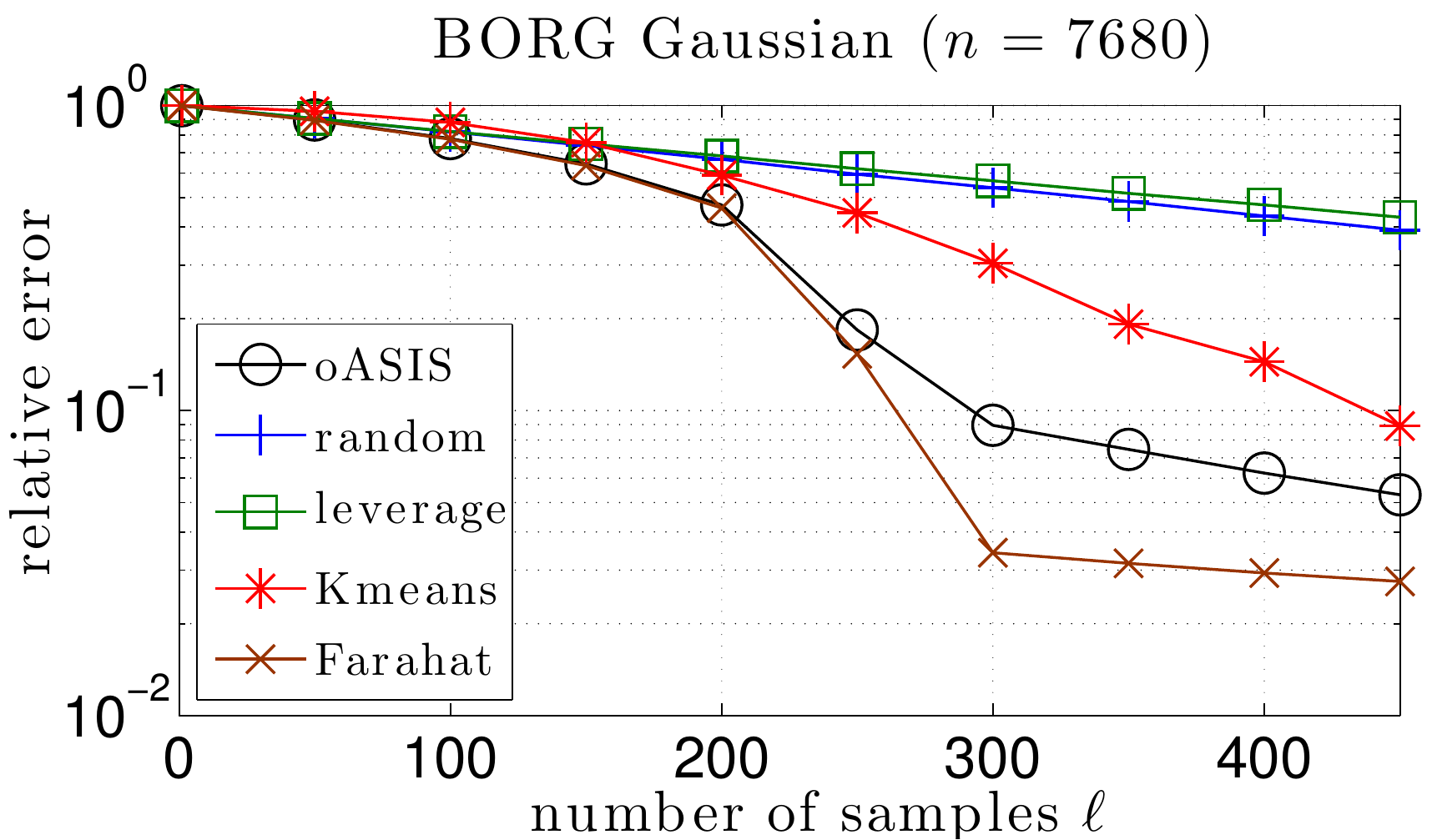}}
\subfloat{%
\includegraphics[scale=0.4, trim = 0mm 0mm 0mm 0mm, clip]{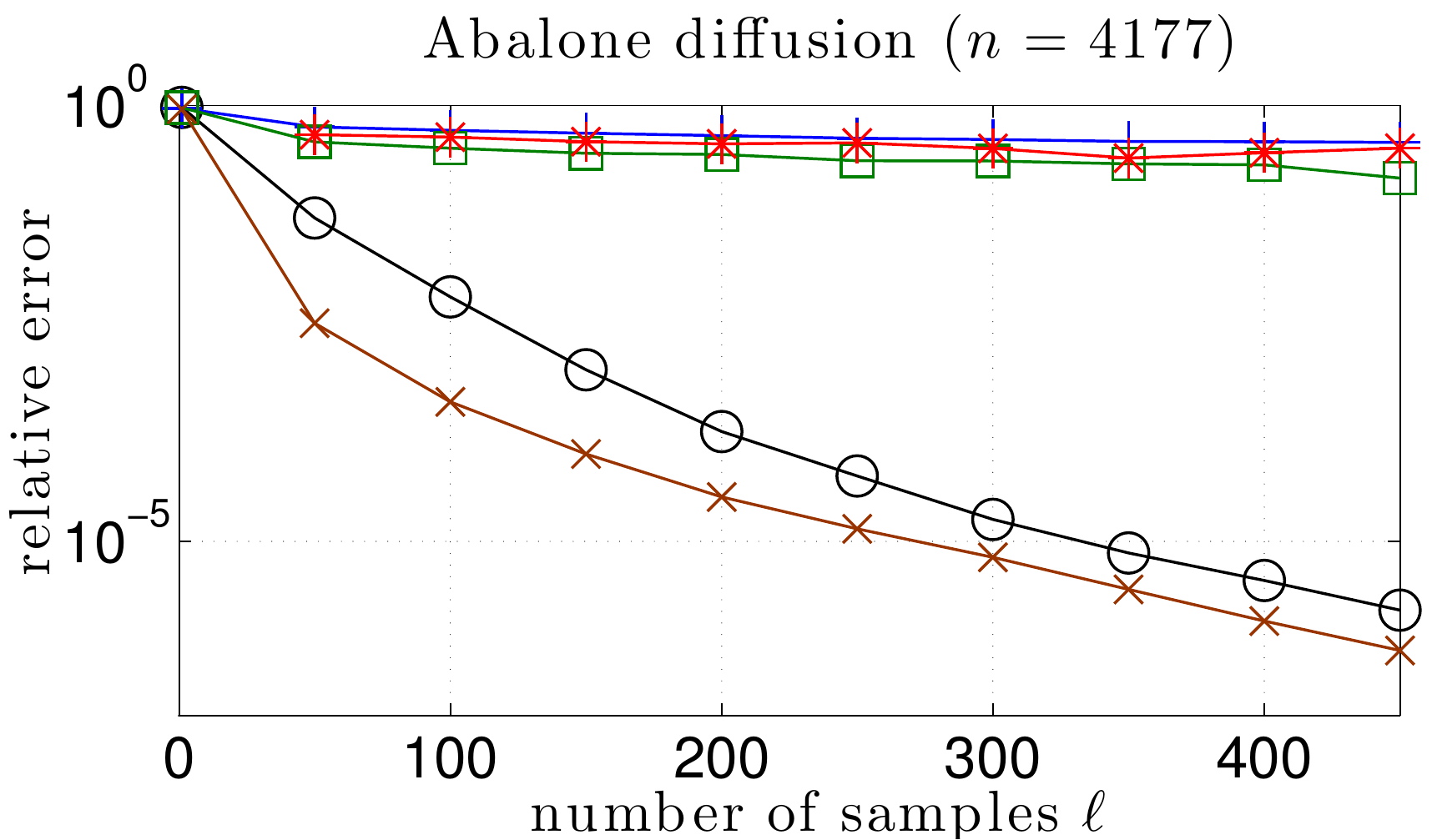}} \quad~
\subfloat{%
\includegraphics[scale=0.4,  trim = 0mm 0mm 0mm 0mm, clip]{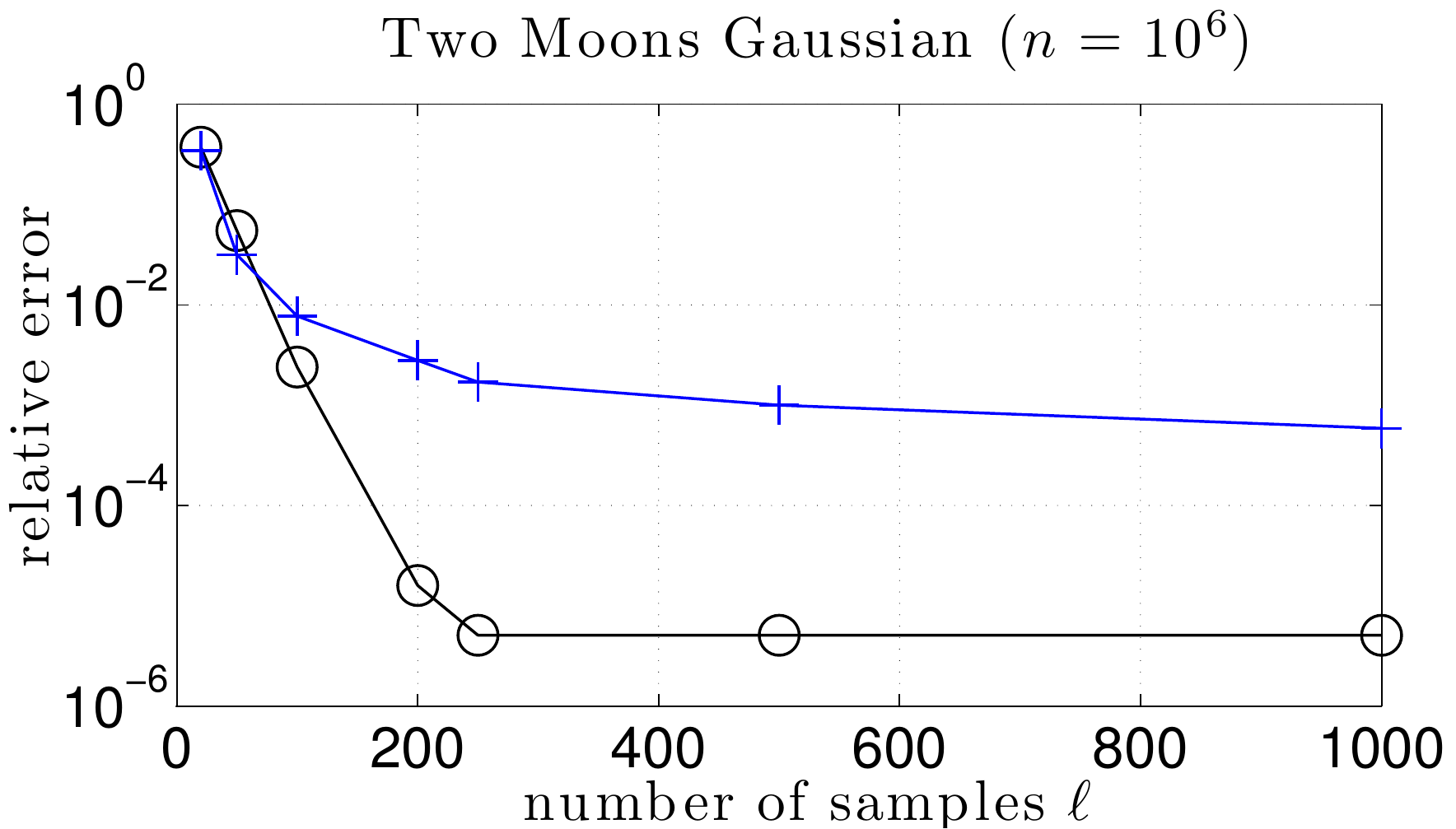}}
\subfloat{%
\includegraphics[scale=0.4,  trim = 0mm 0mm 0mm 0mm, clip]{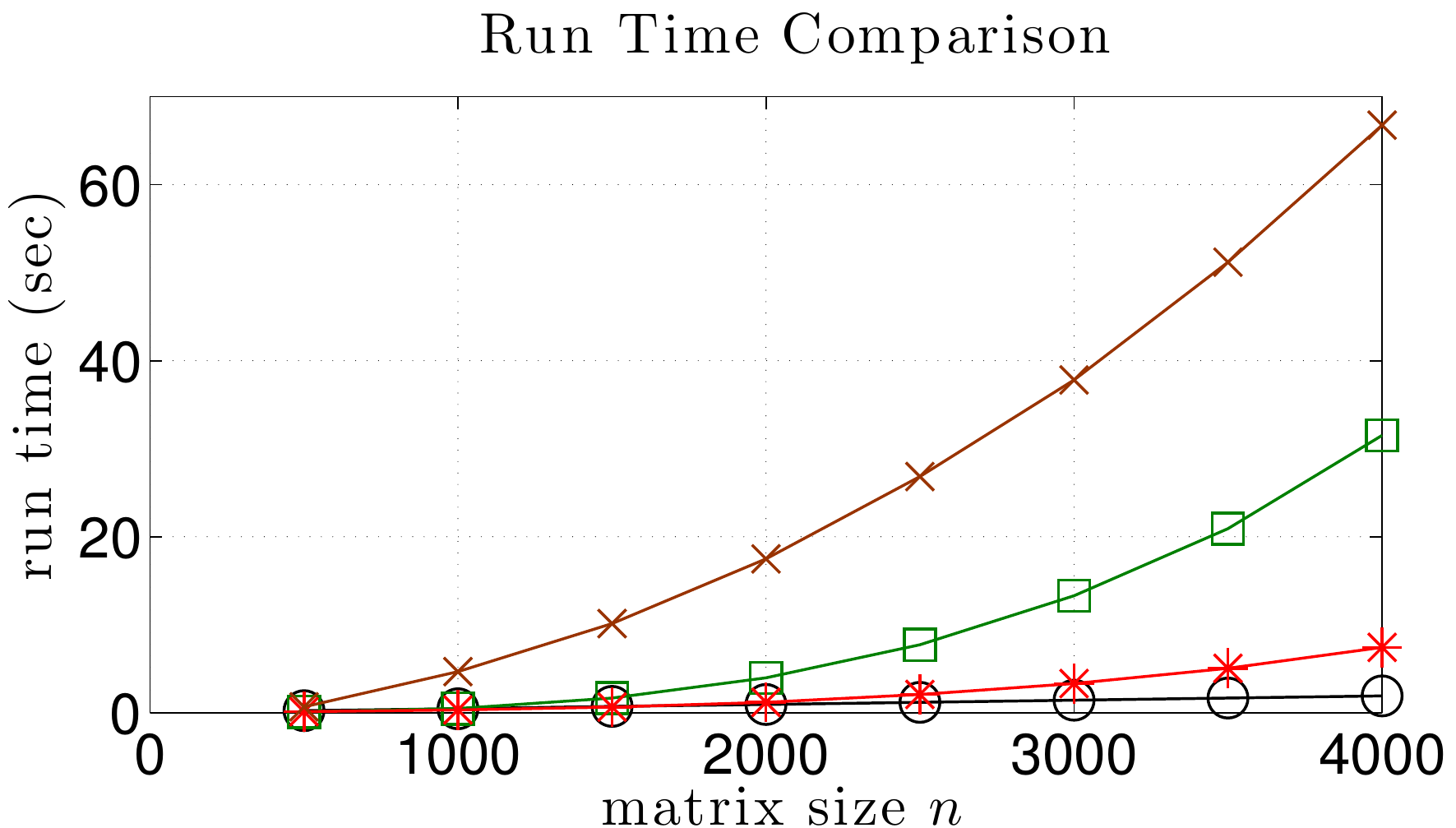}}
\end{center}
\caption{ \nystrom{} Approximation Error Curves and Run Time Comparison. See Sections~\ref{ss:gs} and \ref{ss:sdm} for details. oASIS is accurate and scales well to large problem sizes due to its low runtime complexity (see Section \ref{sec:advantage}).}
\label{fig:BigFigure}
\end{figure*}

\begin{figure*}[!t]
\begin{center}
\subfloat{%
\includegraphics[scale=0.39,  trim = 0mm 0mm 0mm 0mm, clip]{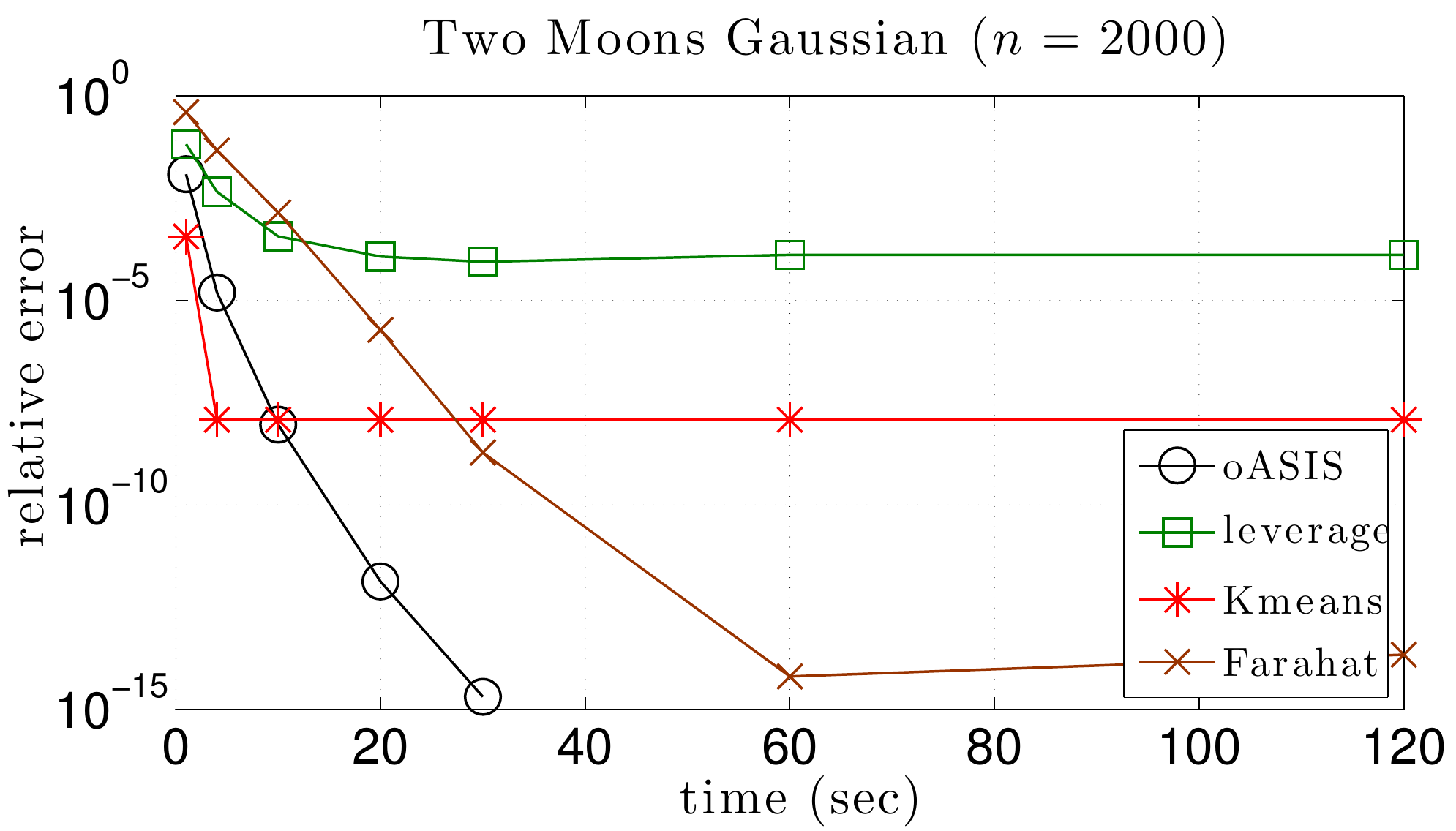}}
\subfloat{%
\includegraphics[scale=0.4,  trim = 0mm 0mm 0mm 0mm, clip]{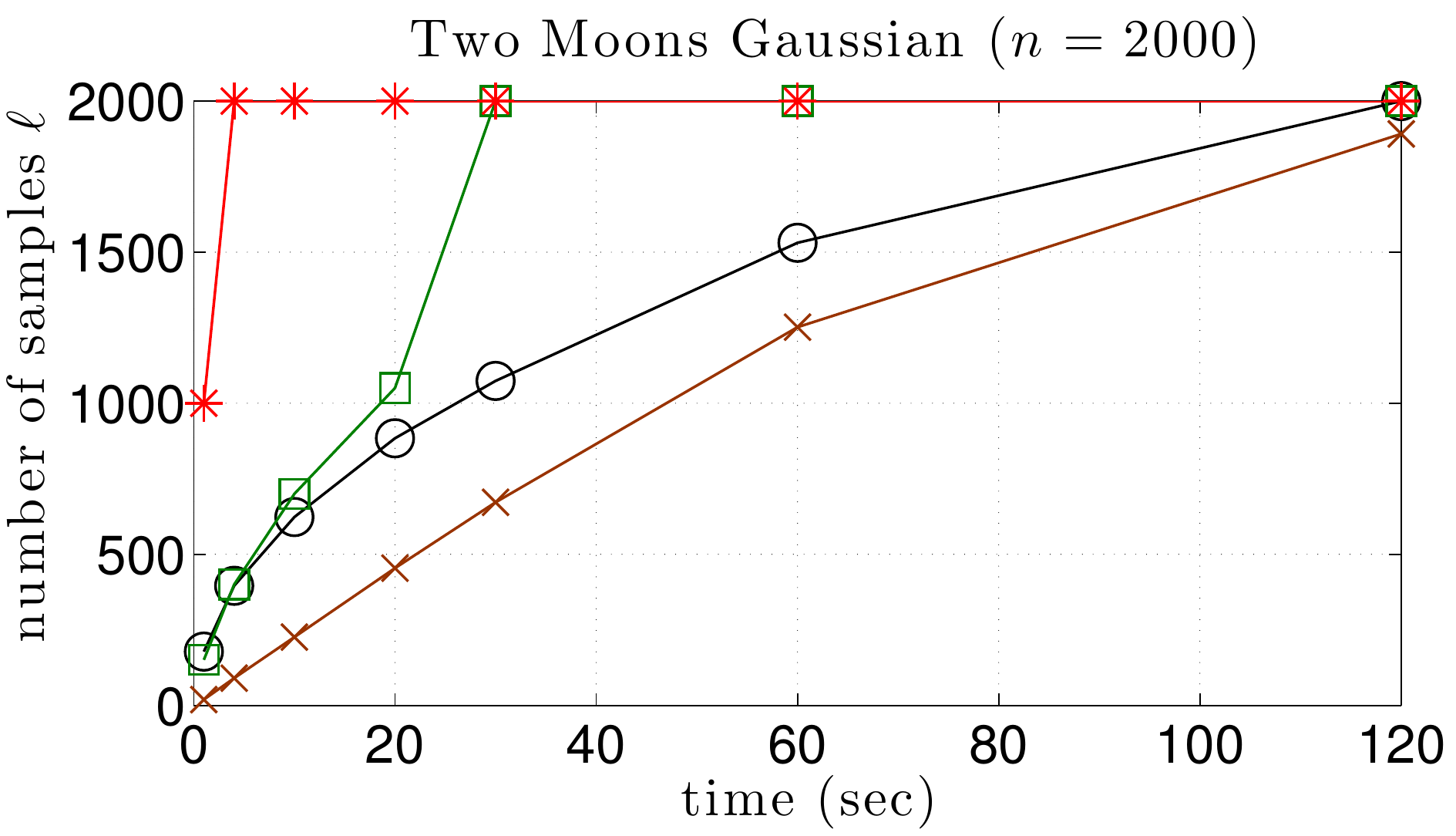}}  \quad~
\subfloat{%
\includegraphics[scale=0.39,  trim = 0mm 0mm 0mm 0mm, clip]{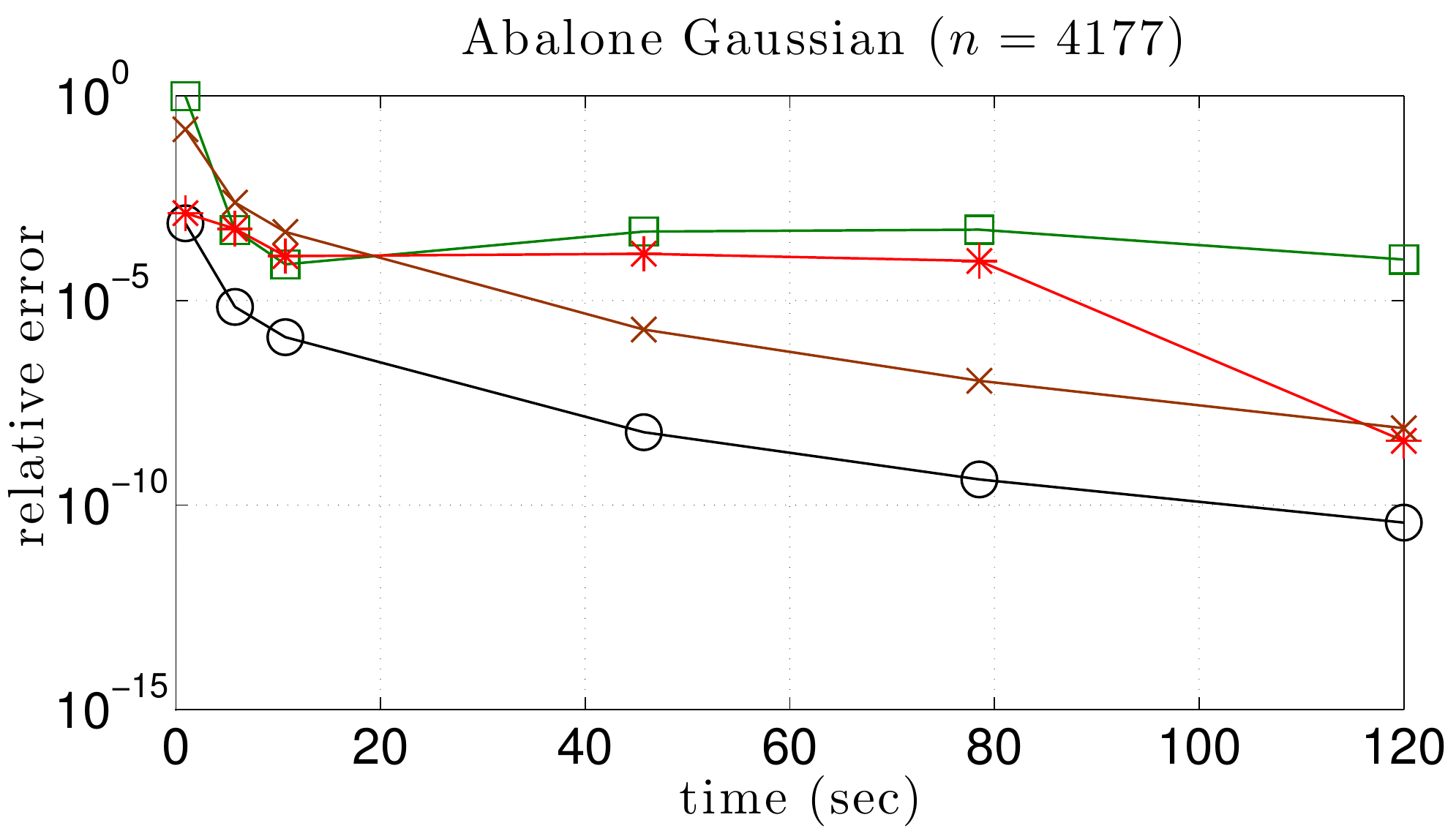}}
\subfloat{%
\includegraphics[scale=0.4,  trim = 0mm 0mm 0mm 0mm, clip]{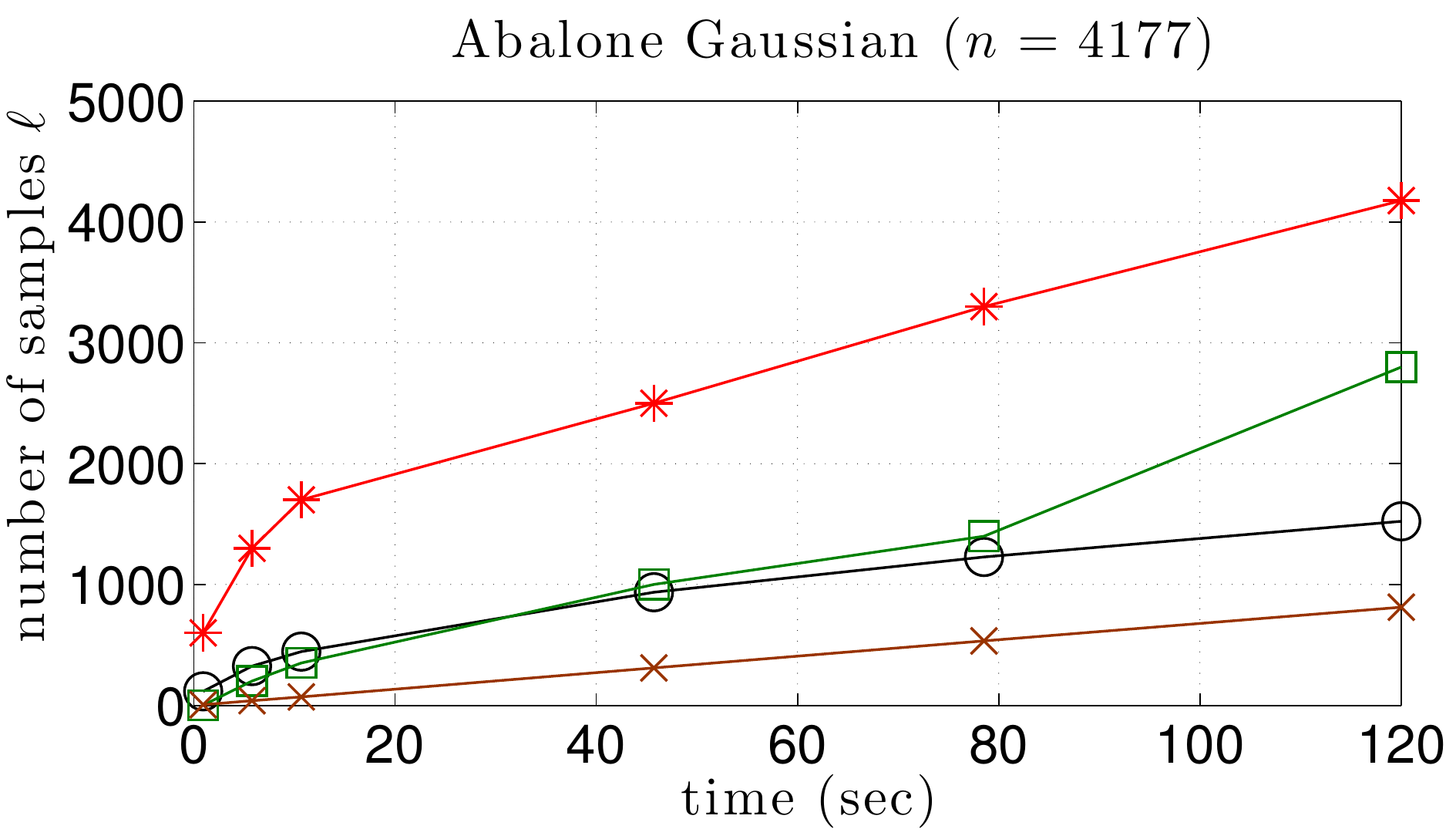}}  \quad~
\subfloat{%
\includegraphics[scale=0.39,  trim = 0mm 0mm 0mm 0mm, clip]{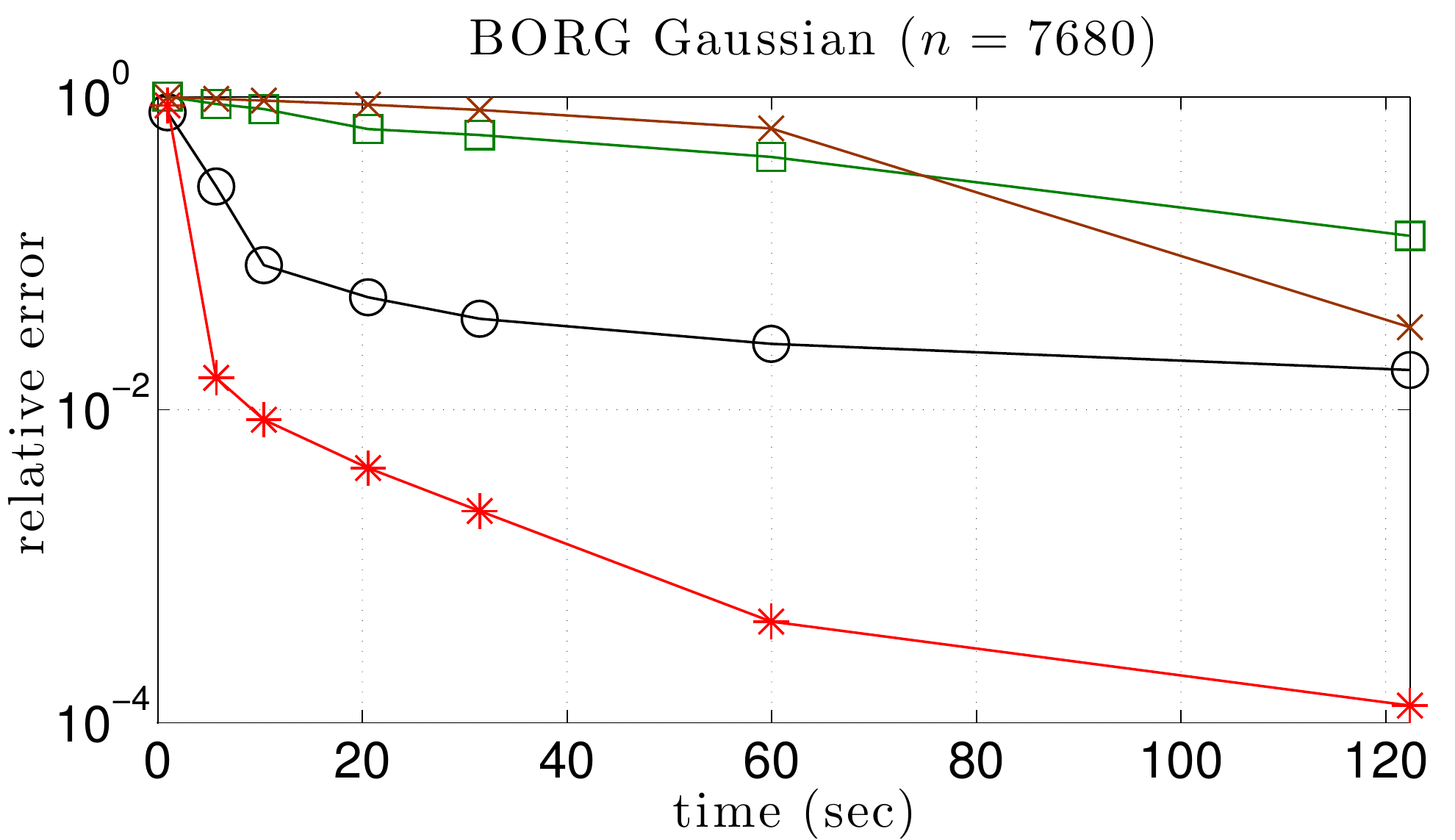}}
\subfloat{%
\includegraphics[scale=0.4,  trim = 0mm 0mm 0mm 0mm, clip]{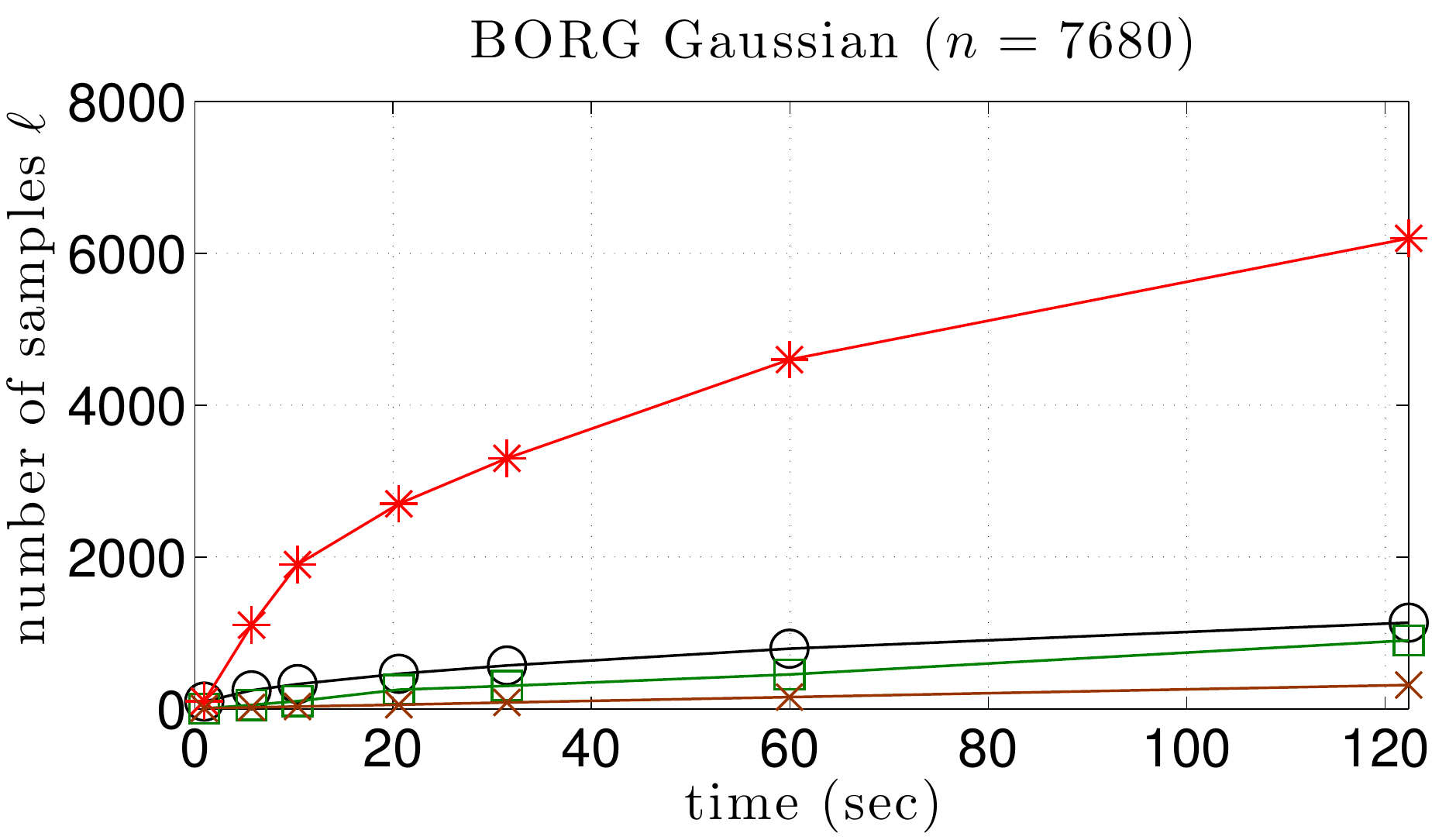}}  \quad~
\end{center}
\caption{ \nystrom{} Error curves and number of samples based on time. See Section~\ref{ss:sdm} for details. Note that oASIS is accurate and scales well to large problem sizes due to its low runtime complexity (see Section \ref{sec:advantage}).}
\label{fig:BigTime}
\end{figure*}

\subsection{General Setup}
\label{ss:gs}

To evaluate the performance of oASIS, we measure the accuracy of \nystrom{} approximations for three size classes of kernel matrices. We first consider matrices where we can directly measure the approximation error of the \nystrom{} method.  Second, we consider larger problems for which forming the entire kernel matrix is impractical. Third, we consider problems so large that the dataset will not fit in memory. For each dataset $\{z_i\}_{i = 1}^n$ in all classes, we consider Gaussian kernel matrices where ${G(i,j) = \exp(\|z_i-z_j\|_2^2/\sigma^2)}$. For datasets in the first class we also consider diffusion distance matrices $M = D^{-1/2}ND^{-1/2}$ where $D$ is a diagonal matrix containing the row sums of $M$, and $N$ is a Gaussian kernel matrix \cite{Coifman20065}. For each dataset, we tune $\sigma$ to provide good matrix approximation for any sampling method.

We compare oASIS to the following state-of-the-art \nystrom{} approximation methods:
(i) uniform random sampling, (ii) Leverage scores \cite{conf/icml/GittensM13} (Section~\ref{sss:ndas}), (iii) $K$-means \nystrom{} approximation \cite{Zhang:2008:INL:1390156.1390311} (Section~\ref{sss:dkm}). and (iv) Farahat's greedy update method \cite{journals/jmlr/FarahatGK11} (Section~\ref{sss:dgm}). For methods (i), (ii), and (iii), we repeat experiments 10 times and average the results. For the second class of matrices we consider oASIS, uniform random sampling, and $K$-means since the other methods become intractable when the matrix becomes too large to explicitly store. For the largest class of matrices we only consider oASIS and uniform random sampling. Specific datasets and experiments are described below.

\subsection{Full Kernel Matrices}
\label{ss:sdm}
Here, we consider datasets for which the kernel matrices can entirely fit in memory, making all the sampling methods tractable.  Convergence curves are generated by forming $\wtg_k$ for increasing $k$ and then calculating the approximation error defined by $\| \widetilde{G}_k-G \|_F /  \|G\|_F.$ We consider the following datasets, run using MATLAB on an iMac with a 2.7 GHz processor and 16GB of memory. Results and column selection runtimes at the largest sample sizes are shown for full matrices in Table~\ref{tab:fulltable}. oASIS is competitive with the most accurate adaptive schemes, at a fraction of the runtime. Figure~\ref{fig:BigFigure} shows selected convergence curves (normalized error vs. \mbox{number} of columns sampled).  Figure~\ref{fig:BigFigure} also presents a plot of column selection runtime vs. \mbox{matrix size} for a variety of methods.  Figure~\ref{fig:BigTime} shows convergence curves (normalized error vs. \mbox{number} of seconds runtime) and column sampling rates (\mbox{number} of columns sampled vs. \mbox{number} of seconds runtime) for all adaptive methods using the Gaussian kernel. These curves allow for a fair assessment of approximation error achieved after a set run time for various adaptive methods, as methods will select columns at different rates. See Section~\ref{ss:discussion} for a full discussion of these results.

\paragraph{Two Moons}
We consider a common synthetic dataset for clustering algorithms that consists of 2-dimensional points arranged in two interlocking moons. This set is 2,000 points of dimension 2. We set the kernel $\sigma$ equal to 5\% of the maximum Euclidean distance between points.

\paragraph{Abalone}
The Abalone dataset is a collection of physical characteristics of abalone \cite{Bache+Lichman:2013}, which are analyzed to find the age of the abalone without direct measurement. This set is 4,177 points of dimension 8. We set the kernel $\sigma$ equal to 5\% of the maximum Euclidean distance between points.

\paragraph{Binary Organization of Random Gaussians (BORG)}
The BORG assimilates sets of points clustered tightly around each vertex of an $n$-dimensional cube. The points around each vertex $v$ are distributed as $\mathcal{N}(v,\sigma_{BORG}^2I)$ with $\sigma_{BORG}^2=0.1$. This dataset is constructed to be pathologically difficult, with many clusters and many points per cluster. Many columns are needed from $G$ to ensure sampling from each cluster. Using an 8 dimensional cube with 30 points at each vertex, the total dataset is 7,680 points of dimension 8. We set the kernel $\sigma$ equal to 12.5\% of the maximum Euclidean distance between points.

\subsection{Implicit Kernel Matrices}
\label{ss:ldm}
Here, we consider datasets for which the resulting kernel matrix would become impractical to explicitly calculate and store. Rather, columns from $G$ are generated ``on the fly'' at the time they are sampled. Since a full representation of $G$ is no longer available, we estimate the approximation error as the Frobenius-norm discrepancy between 100,000 randomly sampled entries in $G$ and the corresponding entries in $\wtg_k$.
Because the Leverage scores \cite{conf/icml/GittensM13} and Farahat \cite{journals/jmlr/FarahatGK11} schemes require a full representation of $G$ (which is intractable for these problem instances), we compare only with uniform random sampling and $K$-means. We consider the following datasets, run using MATLAB on an iMac with a 2.7 GHz processor and 16GB of memory.
Results and column selection runtimes at the largest sample sizes are shown for implicit matrices in Table~\ref{tab:imptable}.

\begin{table*}[!t]
{
\caption{Error and Selection Runtime Results for Explicit Gaussian (first line) and Diffusion (second line) Kernel Matrices.}
\label{tab:fulltable}
\begin{center}
\begin{tabular}{lrrlllll}
\toprule[1.5pt]
Problem     	& 	$n$ 	&$\ell$& \multicolumn{1}{c}{oASIS} 		& \multicolumn{1}{c}{Random} 	& \multicolumn{1}{c}{Leverage scores} 	& \multicolumn{1}{c}{$K$-means} & \multicolumn{1}{c}{Farahat} \\ \midrule
Two Moons	& 2,000 	& 450	&   	$1.00e{-6}$ (1.20)   		&	$2.14e{-3}$ (0.01)       	&   	$9.46e{-4}$ (3.96) 		&	$1.05e{-3}$ (0.38)	 	&	$8.31e{-7}$ (19.7)          \\
                     	&		&		& 	$1.10e{-6}$ (1.16) 		&	$1.22e{-2}$ (0.01)          	&    	$7.45e{-3}$ (4.00)   		& 	$5.49e{-3}$ (1.21)		& 	$1.11e{-6}$ (19.6)              \\
Abalone		& 4,177	& 450	&   	$1.23e{-6}$ (2.60) 		&	$2.65e{-3}$ (0.01)          	&	$5.23e{-4}$ (35.8)          	& 	$1.73e{-3}$ (0.84)		& 	$2.85e{-7}$ (64.8)            \\
                     	&		&		&      	$1.62e{-6}$ (2.51) 		&   	$3.76e{-1}$ (0.01)         	&    	$1.47e{-1}$ (35.9)		& 	$3.24e{-1}$ (8.26)		&	$5.61e{-7}$ (64.7)             \\
BORG		& 7,680 	& 450	&    	$5.30e{-2}$ (4.71)		&     	$3.90e{-1}$ (0.01)            	&       $4.31e{-1}$ (252)	         	& 	$8.89e{-2}$ (2.53)		& 	$2.75e{-2}$ (176)      \\
                     	&		&		&   	$6.29e{-2}$ (4.73) 		&  	$3.90e{-1}$ (0.01)   	    	& 	$4.23e{-1}$ (244)             	& 	$7.70e{-2}$ (48.3)		& 	$2.78e{-2}$ (174)             \\
\bottomrule[1.5pt]
\end{tabular}
\end{center}
}
\end{table*}

\begin{table*}[!t]
{
\caption{Error and Selection Runtime Results for Implicit Kernel Matrices.}
\label{tab:imptable}
\begin{center}
\begin{tabular}{lrrlll}
\toprule[1.5pt]
Problem     	& 	$n$ 	&$\ell$& \multicolumn{1}{c}{oASIS} 		& \multicolumn{1}{c}{Random} & \multicolumn{1}{c}{$K$-means} \\ \midrule
MNIST	  	&50,000	&4,000	&       $1.53e{-6}$ (8260)      	&   	$7.48e{-5}$ (946)       	& 	$7.30e{-7}$ (188)	   \\
Salinas	  	&54,129	&5,000	&       $2.22e{-5}$ (13372)      	&   	$1.61e{-4}$ (819)       	& 	$1.33e{-5}$ (1120)		 \\
Light Field	&82,265	&4,000	&       $7.10e{-6}$ (13600)      	&    	$2.54e{-5}$ (989)   		&	$9.44e{-6}$ (1890)		            \\
\bottomrule[1.5pt]
\end{tabular}
\end{center}
} 
\end{table*}

\begin{table*}[!t]
{
\caption{Error and Factorization Runtime Results for Parallelized Implicit Kernel Matrices.}
\label{tab:ptable}
\begin{center}
\begin{tabular}{lrrll}
\toprule[1.5pt]
Problem     	& 	$n$ 		&$\ell$		& \multicolumn{1}{c}{oASIS-P} 		& \multicolumn{1}{c}{Random}  \\ \midrule
Two Moons	& 1,000,000 &1,000	&	$5.10e{-6}$ (108)		&	$5.90e{-4}$ (279)	 \\
Tiny Images	& 1,000,000 &4,500	&	$2.76e{-4}$ (10672)		&	$2.99e{-6}$ (28225)	\\
Tiny Images	& 4,000,000 &4,500	&	$5.20e{-4}$ (14013)		&	$1.70e{-5}$ (26566)	\\
\bottomrule[1.5pt]
\end{tabular}
\end{center}
} 
\end{table*}

\paragraph{MNIST}
The MNIST dataset consists of handwritten digits used as a benchmark test for classification \cite{lecun95b}. MNIST training data contains 50,000 images of $28\times28=784$ pixels each. Similarity matrices formed from the digits are known to have low-rank structure, because there are only 10 different numerical digits. We set the kernel $\sigma$ equal to 50$\%$ of the maximum Euclidean distance between points.

\paragraph{Salinas}
The Salinas dataset is a hyperspectral image taken in 1998 by the Airborne Visible/Infrared Imaging Spectrometer (AVIRIS) over Salinas Valley, CA. The image is of size $512 \times 217$, over 204 spectral bands, and can be used to classify various areas of crops. Each pixel is assigned to one of 16 classes according to a ground truth image. Classes represent crops such as broccoli or lettuce. We consider all pixels assigned to a nonzero class, for a total number of 54,129 data points. We set the kernel $\sigma$ equal to 10.

\paragraph{Light Field}
Light fields are 4-dimensional datasets describing both the intensity and directionality of light as it travels through a plane. We consider patches taken from the ``chessboard'' dataset of the Stanford Multi-Camera Array \cite{Wilburn:2005:HPI:1073204.1073259}.   The samples are 85,265 vectorized 4-dimensional ``patches,'' each with $4\times 4$ spatial resolution and $5\times 5$ angular resolution for a total dimensionality of 400. We set the kernel $\sigma$ equal to 50\% of the maximum Euclidean distance between points.

\subsection{Implicit Kernel Matrices with oASIS-P}
\label{ss:Pidm}
Finally, we consider datasets that cannot be fit in memory. As such, the dataset is split onto a variety of nodes and the kernel matrix is approximated using oASIS-P as described in Figure~\ref{alg:oasispar}. At this size, we compare only with uniform random sampling. We choose a Gaussian kernel for all datasets in this class.

Since a full representation of $G$ is no longer available, we estimate the approximation error as the Frobenius-norm discrepancy between 100,000 randomly sampled entries in $G$ and the corresponding entries in $\wtg_k$. We consider the following datasets, run using OpenMPI with the Eigen C++ library \cite{eigenweb} over 16 nodes (192 cores) on the DaVinCi cluster at Rice University, with each 2.83 GHz processor core having 4GB of memory. Results at the largest sample sizes are shown for parallelized implicit matrices in Table~\ref{tab:ptable}. These sample sizes were chosen at the limit of available run time on the cluster when using uniform random sampling to both sample and form columns.

\paragraph{Two Moons}
This dataset is as described in Section ~\ref{ss:sdm}, but the number of data points has been increased to 1,000,000 points. Determining a good kernel $\sigma$ from the maximum Euclidean distance among all points becomes intractable, and so we found a kernel $\sigma$ of ${0.5\times\sqrt{3}}$ that provided good approximations for all sampling methods at smaller trial datasets of Two Moons. This kernel $\sigma$ was then used for the full set. For this experiment, oASIS was run to an error tolerance of $1e{-4}$, and random sampling was performed for ${k \in \{20,50,100,200,250,500,1000\}}$ samples.

\paragraph{Millions of Tiny Images}
To show the capability of oASIS to approximate kernel matrices over very large datasets, we select two random subsets of the 80 Million Tiny Images dataset \cite{Torralba:2008:MTI:1444381.1444403}, consisting of 1,000,000 and 4,000,000 $RGB$ images of size $32 \times32$. For reference, storing the full kernel matrix for 4,000,000 tiny images in binary64 would take up 128,000 gigabytes of space. 

We compute the approximation over one color channel of the images as we consider the number of data points to be the prime focus of this experiment. Determining a good kernel $\sigma$ from the maximum Euclidean distance among all points becomes intractable, and so we found a kernel $\sigma$ of $20$ that provided good approximations for all sampling methods at smaller trial datasets of Tiny Images. This kernel $\sigma$ was then used for the full set.

\subsection{Discussion of Results}
\label{ss:discussion}
As shown in in Table~\ref{tab:fulltable}, oASIS achieves lower approximation error than both uniform random sampling and Leverage scores for full kernel matrix experiments when given a set number of columns. In addition, it is competitive with both $K$-means \nystrom{} and Farahat's method in terms of accuracy while having substantially faster run times.

In addition, oASIS's strength as a deterministic method enables oASIS to run for a set length of time, as opposed to a fixed $\ell$. When running the experiments in Figure~\ref{fig:BigTime} for adaptive random schemes, it was not known a priori how many columns either $K$-means or Leverage scores should sample. One must guess at the appropriate $K$ or $\ell$ to use given a certain amount of time. Our experiments found the appropriate parameters through exhaustive search, resetting the clock and increasing $\ell$ for each trial until the time limit was reached.

The primary advantage of oASIS over random selection schemes is its approximation accuracy. In the Abalone example in Figure~\ref{fig:BigFigure}, uniform random sampling does not provide better accuracy as more columns are sampled, while oASIS continues to find columns that can add significantly to the accuracy of the approximation. In Figure~\ref{fig:BigTime}, we observe that oASIS achieves exact matrix recovery with Two Moons at about 30 seconds with around 1000 columns sampled. This is an example of the efficiency of oASIS's column sampling. This efficient accuracy is necessary for the subsequent dimensionality reduction critical to most kernel machine learning tasks. We further observe that the error with $K$-means flattens after 5 seconds. $K$-means \nystrom{} first clusters the original data and computes centroids. It then remaps the data onto the eigenspace of the centroids, and then computes an approximate kernel matrix from this remapping. As such, there is a floor for the approximation based on the accuracy of the eigenspace calculation. This results in a best possible error for $K$-means that can be overcome by oASIS. We observe a similar, flat error result for Leverage scores, as the SVD of the entire $G$ can be performed within the runtime limit. This does not occur when the datasets grow larger.

The primary advantage of oASIS over other adaptive schemes is its efficiency. oASIS saves both time and space. In the runtime results shown in Figure~\ref{fig:BigTime}, we observe fast, accurate approximation of both Two Moons and Abalone with fewer columns sampled. For the BORG dataset, oASIS is second only to $K$-means, which is as expected given that BORG's dataset containing of spherical clusters of equal variance exactly fits the inherent data model that $K$-means clusters. When the data do not fit that model, as in Abalone or Two Moons, we can see the efficiency and accuracy gains of oASIS. We observe in Table~\ref{tab:fulltable} that while $K$-means \nystrom{} runs faster for a single sample size $\ell$, it needs to be run multiple times for consistency. For example, while running a single $K$-means \nystrom{} approximation on Two Moons Gaussian with $n=2000$ and $\ell=450$ takes 0.38 seconds, the 10 runs used for consistency takes 3.8 seconds. Furthermore, $K$-means approximations performed for $\ell$ samples provides no remapping for any samples fewer than $\ell$, nor an index set $\Lam$ of columns for CSS.

The advantages of oASIS-P over uniform random selection become clear in its application. First, oASIS-P can guarantee an invertible $W$. Uniform random sampling can not guarantee that $W$ will be invertible, and so we must calculate $W\dinv$ to compute $\wtg$. Indeed, preliminary experiments frequently showed $W_k$ to be rank-deficient. This is most likely due to the ``birthday problem'' - as more columns are selected, the chances that any two columns are of the same direction grows surprisingly fast. As oASIS can iteratively compute $W\inv$, it does not need to invert an $\ell \times \ell$ matrix as a separate step. Note that 1\% of a 1,000,000 point dataset still results in a $10,000 \times 10,000$ $W$ matrix. Each of DaVinCi's cores had 4GB of memory, and uniform random sampling became infeasible after approximately 4,500 columns as computing $W\dinv$ became too memory intensive for a single node, and distributing the computation of $W\dinv$ is not straightforward. Second, while the complexity of oASIS-P appears much higher than uniform random sampling, in practice oASIS-P is faster than uniform random sampling at sampling and forming $\ell$ columns.   Column generation takes the same amount of time regardless of the column selection scheme, and in large data regimes communication of data vectors between nodes becomes the computational bottleneck. Computing an iterative $W\inv$ is faster than computing $W\dagger$, and so for very large data regimes oASIS-P becomes faster than uniform random sampling, as shown in Table~\ref{tab:ptable}. For example, oASIS-P samples and forms columns over the Two Moons dataset in less than half of the time it takes for uniform random sampling to perform the same task.

In addition to its low runtime complexity, oASIS is also capable of benefiting from sparse matrix structure.  For such matrices, adaptive methods like the one in \cite{journals/jmlr/FarahatGK11} requires the computation of $n\times n $ ``residuals,'' which may be dense even in the case that $G$ is extremely sparse.  In contrast, oASIS requires only the storage of much smaller $\ell\times n$ matrices.  This benefit of oASIS is highly relevant for extremely large datasets where sparse approximations to similarity matrices are formed using $K$-nearest-neighbor algorithms that only store the most significant entries in each matrix column. Further analysis is necessary, however, as fast approximation methods have been developed specifically for sparse kernel matrices \cite{Halko:doi:10.1137/090771806}.

\section{Conclusion}
In this paper, we have introduced oASIS, a novel adaptive sampling algorithm for \nystrom{} based low-rank approximation. oASIS combines the high accuracy of adaptive matrix approximation with the low runtime complexity and memory requirements of inexpensive random sampling. oASIS achieves exact matrix recovery in an optimal number of columns. We have demonstrated the speed and efficacy of the method by accurately approximating large matrices using a single processor. In addition, we have parallelized oASIS so it could be run over datasets of arbitrary size. This allows oASIS to be the only adaptive greedy method available in large data regimes. In addition, our numerical experiments show oASIS to be competitive with random schemes at this level, in both accuracy and speed. Using a dataset of 1,000,000 examples we are able to achieve 1\% of the approximation error of random sampling methods. oASIS has been applied to sparse subspace clustering \cite{Dyer2015SEED}, and future work will focus on applying oASIS to other machine learning tasks, such as manifold learning and spectral clustering. 

\bibliographystyle{IEEEtran}
\bibliography{IEEEabrv,oasis_submission}

\end{document}